\documentclass[10pt,twocolumn,letterpaper]{article}

\usepackage{iccv}
\usepackage{times}
\usepackage{epsfig}
\usepackage{graphicx}
\usepackage{amsmath}
\usepackage{amssymb}
\usepackage{booktabs}
\usepackage{enumitem}
\usepackage[numbers,sort]{natbib}
\usepackage[font=small,labelfont=bf]{caption}
\usepackage{multirow}

\newcommand{\norm}[1]{\left\lVert#1\right\rVert}
\newcommand{\addpic}[1]{\includegraphics[width=8em]{#1}}
\newcommand{\addpicBig}[1]{\includegraphics[width=24em]{#1}}
\usepackage[breaklinks=true,bookmarks=false,colorlinks=true,citecolor=green]{hyperref}

\iccvfinalcopy 

\def\iccvPaperID{3758} 

\ificcvfinal\fi
\begin{document}

\title{Canonical Surface Mapping via Geometric Cycle Consistency}



\author{Nilesh Kulkarni \qquad \qquad Abhinav Gupta\textsuperscript{*} \qquad \qquad Shubham Tulsiani\textsuperscript{*} \\
Carnegie Mellon University \qquad Facebook AI Research\\
{\tt \small \{nileshk, abhinavg\}@cs.cmu.edu} \qquad   \tt \small shubhtuls@fb.com
\\ {\tt \small \href{https://nileshkulkarni.github.io/csm/}{https://nileshkulkarni.github.io/csm/}}
}

\twocolumn[{%
\renewcommand\twocolumn[1][]{#1}%
\vspace{-3em}
\maketitle
\vspace{-2em}
\begin{center}
   \centering \includegraphics[width=\textwidth]{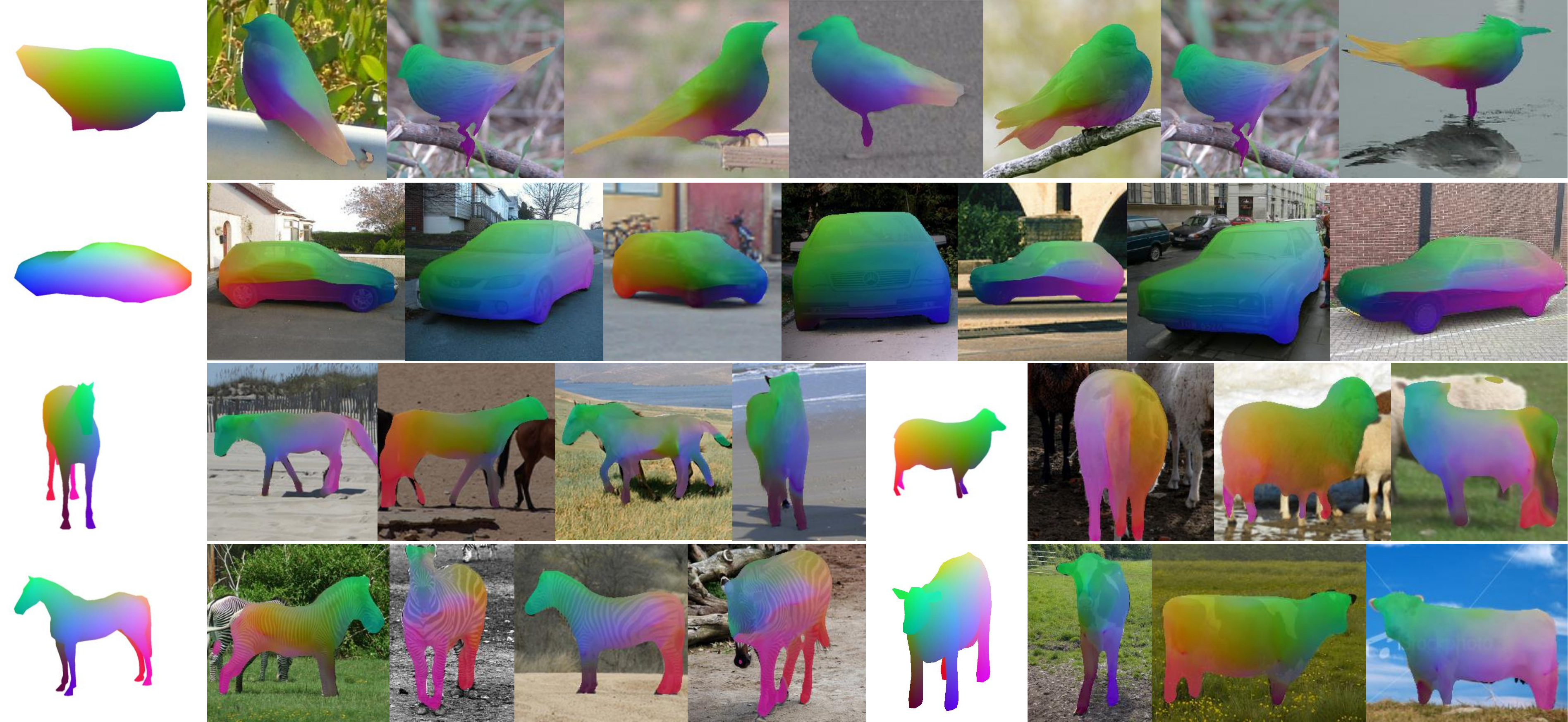} \captionof{figure}{We study the task of Canonical Surface Mapping (CSM). This task is a generalization of keypoint estimation and involves mapping pixels to canonical 3D models. We learn CSM prediction without requiring correspondence annotations, by instead using geometric cycle consistency as supervision. This allows us to train CSM prediction for diverse classes, including rigid and non-rigid objects.}
   \label{fig:teaser}
\end{center}%
}]

\thispagestyle{empty}

\begin{abstract}
We explore the task of Canonical Surface Mapping (CSM).  Specifically, given an image, we learn to map pixels on the object to their corresponding locations on an abstract 3D model of the category.
But how do we learn such a mapping? A supervised approach would require extensive manual labeling which is not scalable beyond a few hand-picked categories. Our key insight is that the CSM task (pixel to 3D), when combined with 3D projection (3D to pixel), completes a cycle. Hence, we can exploit a geometric cycle consistency loss, thereby allowing us to forgo the dense manual supervision. Our approach allows us to train a CSM model for a diverse set of classes, without sparse or dense keypoint annotation, by leveraging only foreground mask labels for training. We show that our predictions also allow us to infer dense correspondence between two images, and compare the performance of our approach against several methods that predict correspondence by leveraging varying amount of supervision.
\end{abstract}
\vspace{-4mm}
\section{Introduction}
Plato famously remarked that while there are many cups in the world, there is only one `idea' of a cup. Any particular instance of a category can thus be understood via its relationship to this platonic ideal. As an illustration, consider an image of a bird in 
Figure~\ref{fig:teaser}. When we humans see this image, we can not only identify and segment the bird but also go further and even map pixels to an abstract 3D representation of the category.
This task of mapping pixels in an image to locations on an abstract 3D model (which we henceforth call {\bf canonical surface mapping}) is generalization and densification of keypoint estimation and is key towards rich understanding of objects. But how do we learn to do this task? What is the right data, supervision or models to achieve dense rich understanding of objects?
\let\thefootnote\relax\footnotetext{* the last two authors were equally uninvolved.}

One way to learn the canonical surface mapping task is to collect large-scale labeled data. Specifically, we can label hundreds or thousands of keypoints per image for thousands of images. As each keypoint location defines which pixel corresponds to a specific location on the 3D surface, this approach of manually labeling the keypoints can provide dense supervision for learning. 
This approach has in fact been shown to be quite successful for specific categories such as humans~\cite{alp2018densepose}. But of course collecting such labeled data requires enormous manual labeling effort, making it difficult to scale to generic categories.  


Is there an alternative supervisory signal that can allow one to learn without reliance on such labelled data? Interestingly, we note that this task of canonical surface mapping is an inverse graphics task. Any such mapping is constrained by the geometry operating on the underlying 3D, and any predicted mapping should also respect this structure. In particular, for the pixels that belong to the object given by the object mask, the CSM function maps these pixels onto the 3D shape. These points on the 3D shape, when projected back using (known/predicted) camera, should map back to the same pixels. Our key insight is that one can complete the cycle (pixels $\rightarrow$ 3D $\rightarrow$ pixels) and use the consistency loss as an objective. The gradients from the loss can be propagated back to the CSM function prediction function, thereby allowing us to learn this mapping without reliance on strong forms of supervision.

In this paper, we present an approach to learn the task of canonical surface mapping from the set of images belonging to semantic category, their input masks and an abstract 3D model which represents the semantic category. Additionally, we show that predicting a canonical surface mapping for images allows us to infer dense correspondence across images of a category, and our approach enables recovery of dense correspondences without any correspondence supervision! In comparison to approaches that use dense supervision for this task~\cite{alp2018densepose}, or approaches that leverage keypoints for the related tasks of semantic correspondence~\cite{choy2016universal}, or 3D reconstruction~\cite{cmrKanazawa18}, this is significant decrease in supervision. This allows us to train our CSM model for a diverse set of classes: birds, zebras, cars and more (See Figure~\ref{fig:teaser}). We believe our approach can pave the way for large-scale internet-driven 3D understanding and correspondence inference since both semantic imagesets and masks are easy to obtain (and automatic approaches can be used as well). 

\vspace{-1mm}
\section{Related Work}
\noindent \textbf{Dense Semantic Correspondences.}
A fundamental task that is equivalent to pursuing canonical surface mapping is that of inferring dense semantic correspondence -- given two images, the goal is to predict for each pixel in the former, the corresponding pixel in the latter.
Methods prior to the recent resurgence of deep learning~\cite{liu2011sift,kim2013deformable} demonstrated that matching using features such as SIFT could allow recovering  correspondence across instances, and later work showed similar results using CNN features~\cite{long2014convnets,ham2016proposal}. While these generic features allow recovering correspondence, learning specifically for the task using annotated data can improve results~\cite{choy2016universal}. However, collecting such annotation can be tedious, so several approaches have attempted to relax the supervision for learning correspondence.

Among these, a common paradigm is to learn correspondence by self-supervision, where random perturbations of images are used as training pairs. This allows predicting parametric warping~\cite{kanazawa2016warpnet,Rocco_2017_CVPR,rocco2018end} to relate images, or learn equivariant embeddings~\cite{thewlis2017unsupervised} for matching. However, the these methods are fundamentally restricted to training data of the same instance, with no change in the visible content, thereby limiting the performance for different instances with viewpoint changes.  While for certain categories of interest \eg humans, some approaches \cite{taylor2012vitruvian,tung2017self,pons2015metric, rhodin2018unsupervised, simon2017hand} show that it is possible to use calibrated multi-view or motion capture to generate supervision, this form of supervision is slightly tedious to collect for all classes. 
An alternate form of supervision can come via synthetic data, where synthetic image pairs rendered using the same pose as a real image pair, can help learn a correspondence function between real images that is cycle-consistent~\cite{zhou2016learning}. However, this approach relies on availability of large-scale synthetic data and known pose for real images to generate the supervisory signal, and we show that both these requirements can be relaxed.

\vspace{2mm}
\noindent \textbf{Learning Invariant Representations.}
Our work is broadly related to methods that learn pixel embeddings invariant to certain transforms. These approaches leverage tracking to obtain correspondence labels, and learn representations invariant to viewpoint transformation~\cite{schmidt2017self,zeng20173dmatch} or motion~\cite{Wang_UnsupICCV2017}. Similar to self-supervised correspondence approaches, these are also limited to training using observations of the same instance, and do not generalize well across instances. While our canonical surface mapping is also a pixel-wise embedding invariant to certain transforms, it has a specific geometric meaning \ie correspondence to a 3D surface, and leveraging this is what allows learning without the correspondence supervision.




\vspace{2mm}
\noindent \textbf{Category-Specific 3D Reconstruction.}
A related line of work pursued in the community is that of reconstructing the instances in a category using using category-specific deformable models. Dating back to the seminal work of Blanz \& Vetter~\cite{BlanzVetter}, who operationalized D'Arcy Thompson's insights into the manifold of forms~\cite{Thompson}, morphable 3D models have been used to model faces~\cite{BlanzVetter}, hands~\cite{taylor2014user,khamis2015learning}, humans~\cite{Anguelov:2005,SMPL} and other generic classes~\cite{cashman2013shape,CSDM,3dinterpreter,cmrKanazawa18}.

In conjunction with known/predicted camera parameters, this representation also allows one to extract a pixelwise canonical mapping.
However, these methods often rely on 3D training data to infer this representation. Even approaches that relax this supervision~\cite{CSDM,3dinterpreter,cmrKanazawa18} crucially rely on (sparse or dense) 2D keypoint annotations during training. In contrast, we show that learning a canonical surface mapping is feasible even without such supervision. Further, we demonstrate that directly learning the mapping function leads to more accurate results than obtaining these via an intermediate 3D estimate.

\vspace{2mm}
\noindent \textbf{Consistency as Meta-Supervision.}
Ours is not the only task where acquiring direct supervision is often infeasible, and the idea of leveraging some form of consistency to overcome this hurdle has been explored in several domains. Recent volumetric reconstruction ~\cite{rezende2016unsupervised,yan2016perspective,drcTulsiani17,gwak2017weakly} or depth prediction~\cite{garg2016unsupervised,zhou2017unsupervised,godard2017unsupervised} approaches use geometric consistency between the predicted 3D and available views as supervision.
Similarly, the notion that when learning some transformations, their composition often respects a cyclical structure has been used for image generation~\cite{zhu2017unpaired,kim2017learning}, correspondence estimation~\cite{zhou2015flowweb,zhou2017unsupervised} \etc. In our setup, we also observe that the approach of using consistency as meta-supervision allows bypassing supervision. We do so by leveraging insights related to both, geometry and cycle consistency -- given a surface mapping, there is a geometrically defined inverse transform with which the canonical surface mapping predictions should be cycle-consistent.

\vspace{-1mm}
\section{Approach}
Given an image, our goal is to infer for each pixel on the object, its mapping onto a given canonical template shape of the category. We do so by learning a parametrized CNN $f_{\theta}$, which predicts a pixelwise canonical surface mapping (CSM) given an input image. We show that our method, while only relying on foreground masks as supervision, can learn to map pixels to the given category-level template shape. Our key insight is that this mapping function we aim to learn has a geometric structure that should be respected by the predictions. We operationalize this insight, and learn a CSM predictor using a geometric cycle consistency loss, thereby allowing us to bypass the need for supervision in the form of annotated (sparse or dense) keypoints.

We first present in Section \ref{subsec:known-cam} our training setup in a scenario where the camera pose for each training image is given. We then show how we can relax this requirement of known camera in Section \ref{subsec:relax-cam}. Learning a CSM predictor implicitly allows us to capture the correspondence across instances, and we describe in Section \ref{subsec:correspondence} the procedure to recover dense semantic correspondence given two images.

\subsection{Preliminaries}
\noindent
{\bf Surface Parametrization.} 
The template shapes we learn mappings to are in fact two-dimensional surfaces in 3D space. The surface $S$ of the template shape can therefore be parametrized via two parameters $u \in (0,1)$ and $v \in (0,1)$ (or equivalently a 2D vector $\mathbf{u}$). This parametrization implies that we can obtain a mapping $\phi$ such that $\phi(\mathbf{u})$ represents a unique point on the surface $S$.

\begin{figure}[h!]
  \centering
   \includegraphics[width=0.4\textwidth]{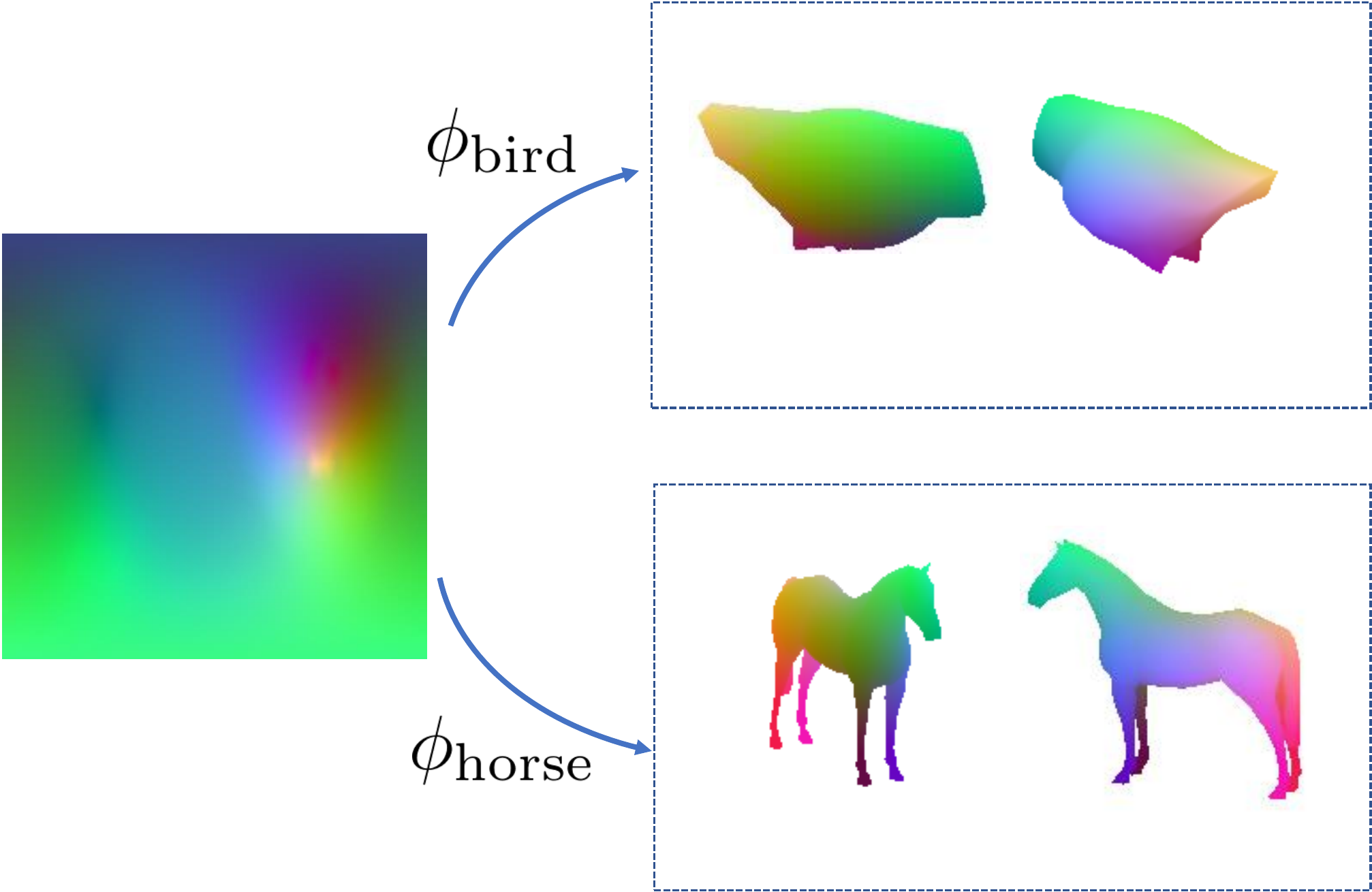}
  \caption{{\bf Surface Parametrization.} We show the mapping from $(u,v)$ space to the surface of the 3D model for two categories.}
  \label{fig:surfaceuv}
\end{figure}

While there are several ways to construct such a mapping, one intuitive way is to consider $\mathbf{u}$ to represent the polar angles to parametrize points on the surface of a hollow sphere, which can be mapped to a surface $S$ by pushing it inward \cite{praun2003spherical}. Given a template shape with a surface $S$, we use this approach to obtain the parametrization $\phi$. We show some visualizations in Figure \ref{fig:surfaceuv} for the mapping from a 2D square to template 3D shapes for two categories.

\vspace{2mm}
\noindent {\bf Canonical Surface Mapping.}
A canonical surface mapping $C$ for an image $I$ is a mapping from pixels onto the template 3D shape. Given a pixel $\mathbf{p} \equiv (x,y)$, $C[\mathbf{p}]$ represents the corresponding point on the surface. As the surface has a two-dimensional parametrization, $C$ is equivalently an image of the same size as $I$, with a two-channel value at each pixel. Our parametrized CNN $f_{\theta}$ that predicts this mapping from an input image, therefore learns a per-pixel prediction task -- given an RGB input image, it outputs a 2 dimensional vector for each pixel.

\vspace{2mm}
\noindent {\bf Camera Projection.}
We model the camera as a weak perspective (scaled orthographic) transformation. We represent the camera for every image $I$ as $\pi$, parameterized by the scale  $s \in \mathcal{R}$, translation $t \in \mathcal{R}^2$ and rotation $r$ are three euler angles. We denote by $\pi(P)$ as the projection of a point $P$ to the image coordinate frame using the camera parameters $\pi \equiv (s,t,r)$.

\subsection{Learning via Geometric Cycle Consistency}
\label{subsec:known-cam}
\noindent
We aim to learn a per-pixel predictor $f_{\theta}$ that outputs a canonical surface mapping given an input image $I$. We present an approach to do so using only foreground masks as supervision. However, for simplicity, we first describe here how we can learn this CSM predictor assuming known camera parameters for each training image, and relax this requirement in Section \ref{subsec:relax-cam}. 

\begin{figure}[h!]
\centering
\includegraphics[width=0.45\textwidth]{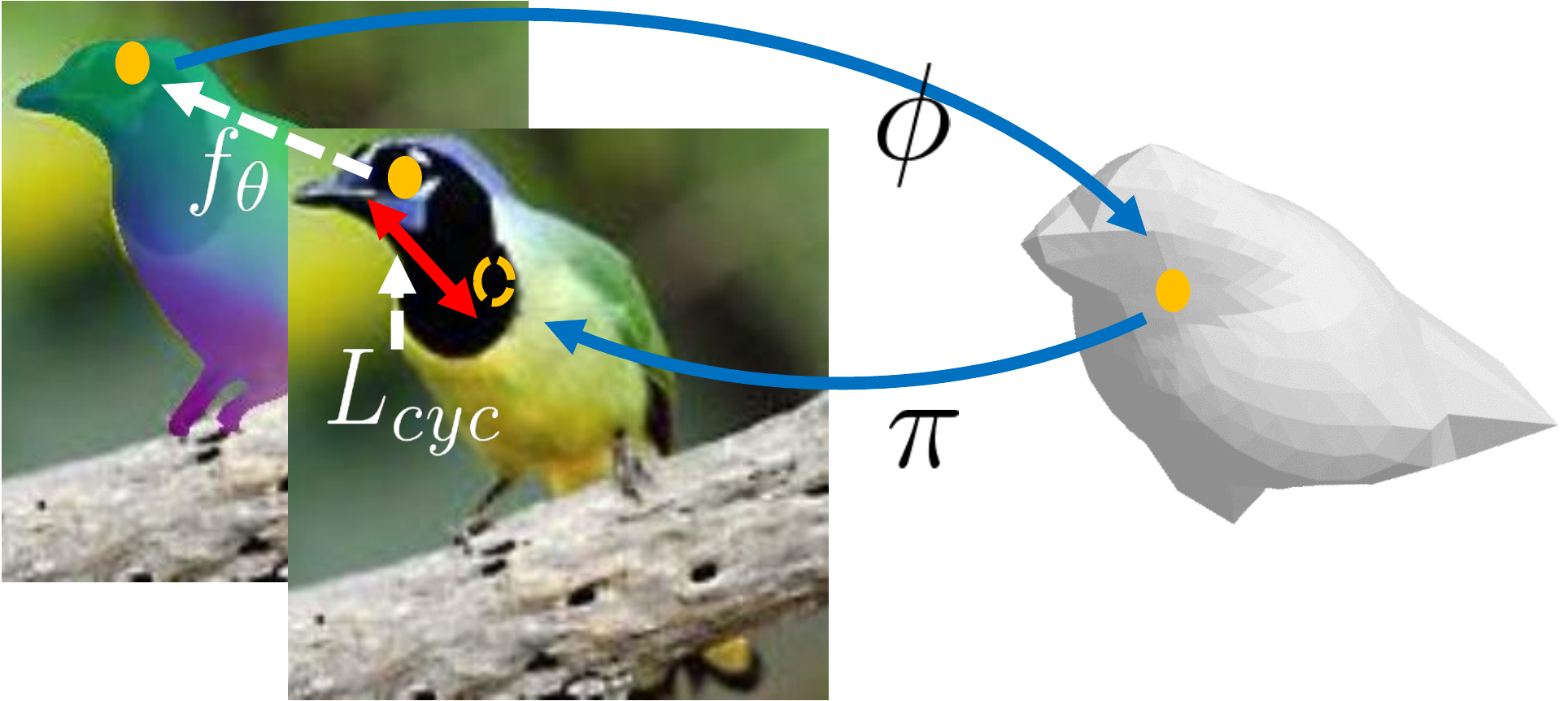}
  \caption{{\bf Geometric Cycle Consistency Loss.}
  A pixel mapped to $\mathbf{u}$ by CSM function $f_\theta$ gets mapped onto the 3D template via $\phi$. Our loss enforces that  this 3D point, when projected back via the camera $\pi$, should map back to the pixel.}
  \label{fig:gccloss}
\end{figure}

Our approach is to derive learning signal from the geometric nature of this task. In particular, as the 3D shapes underlying instances of a category are often similar (and therefore similar to the template shape), a pixel-wise mapping onto the 3D surface should be (approximately) cycle-consistent under reprojection. We capture this constraint via a geometric cycle consistency loss. This loss, in conjunction with an objective that allows the prediction to respect certain visibility constraints, allows us to learn $f_{\theta}$.





\vspace{2mm}
\noindent {\bf Geometric Cycle Consistency Loss.} Given an image $I$ with associated camera $\pi$ and foreground mask $I_f$, we wish to enforce that the predicted canonical surface mapping $C \equiv f_{\theta}(I)$, respects the underlying geometric structure. Concretely, as the instances across a category bear resemblance to the template shape, given a pixel $\mathbf{p}$ on the object foreground, we would expect that its corresponding point on the 3D surface $\phi(C [\mathbf{p}])$ to (approximately) project back under the camera $\pi$ which we denote as $\mathbf{\bar{p}}$. We define a geometric consistency loss (see Figure \ref{fig:gccloss}) that penalizes this inconsistency for all foreground pixels, thereby encouraging the network to learn pixel $\rightarrow$ 3D mapping functions that are cycle-consistent under the 3D $\rightarrow$ pixel reprojection.
\begin{align}
L_{\text{cyc}} = \sum_{\mathbf{p} \in I_{f}} \| \mathbf{\bar{p}} - \mathbf{p} \|^2_2 \quad ; \quad \mathbf{\bar{p}} = \pi(\phi(C[\mathbf{p}]))
\end{align}
\noindent
{\bf Incorporating Visibility Constraints.} Enforcing that the pixels when lifted to 3D, project back to the same location is desirable, but not a sufficient condition. As an illustration, for a front facing bird, both the beak and tail project at similar locations, but only the former would be visible. This implies that points on the surface that are self-occluded under $\pi$ can also result in minimizing $L_{cyc}$. Our solution is to discourage $f_\theta$ from predicting $\mathbf{u}$ values that map to self-occluded regions under camera $\pi$.

A point on the 3D shape is self-occluded under a camera $\pi$, its z-coordinate in camera frame is larger than the rendered depth at the corresponding pixel. We use Neural Mesh Renderer (NMR)~\cite{kato2018neural} to render a depth map $D_{\pi}$ for the template shape $S$ under camera $\pi$, and define a visibility loss for each pixel $\mathbf{p}$ by checking if the z-coordinate (say $z_{\mathbf{p}}$) of its corresponding point $\phi(C [\mathbf{p}])$ on the 3D shape, when projected under $\pi$, has a larger z-coordinate.




\begin{align}
    L_{\text{vis}} = \sum_{p \in I_{f}}\max(0,z_{\mathbf{p}} - D_{\pi}[\mathbf{\bar{p}}])
\end{align}

\vspace{2mm}
\noindent {\bf Network Details.}
We implement $f_{\theta}$ as a network with UNet~\cite{ronneberger2015u} style architecture.  This network takes as input an image of size 256 x 256 and outputs a unit vector per pixel representing a point on surface of sphere which is then converted to a $(u,v)$ coordinate analogous to latitude and longitude. We train our network to minimize the cycle-consistency and visibility objectives:
\begin{align}
    L_{\text{consistency}} = L_{\text{vis}} + L_{\text{cyc}}
\end{align}

\noindent
Even though we do not have direct supervision for the mappings, as we train a shared predictor across instances, the explicit priors for geometric consistency, and the implicit inductive biases in CNNs for spatial equivariance are sufficient for us to learn a meaningful predictor.

\vspace{2mm}
\noindent {\bf Foreground Mask Prediction.}
While the training procedure described above encourages cycle-consistent predictions at pixels belonging to the object, the learned CNN $f_{\theta}$ also predicts some (possibly spurious) values at other pixels. To allow us to ignore these background pixels for inferring correspondence (see Section. \ref{subsec:correspondence}), as well as for generating visualizations, we train an additional per-pixel mask predictor using standard cross-entropy loss $L_{fg}$ against the ground-truth masks. To do so, we simply modify $f_{\theta}$ to yield an additional per-pixel foreground probability as output.

\begin{figure}[h]
\centering
    \centering
\includegraphics[width=0.45\textwidth]{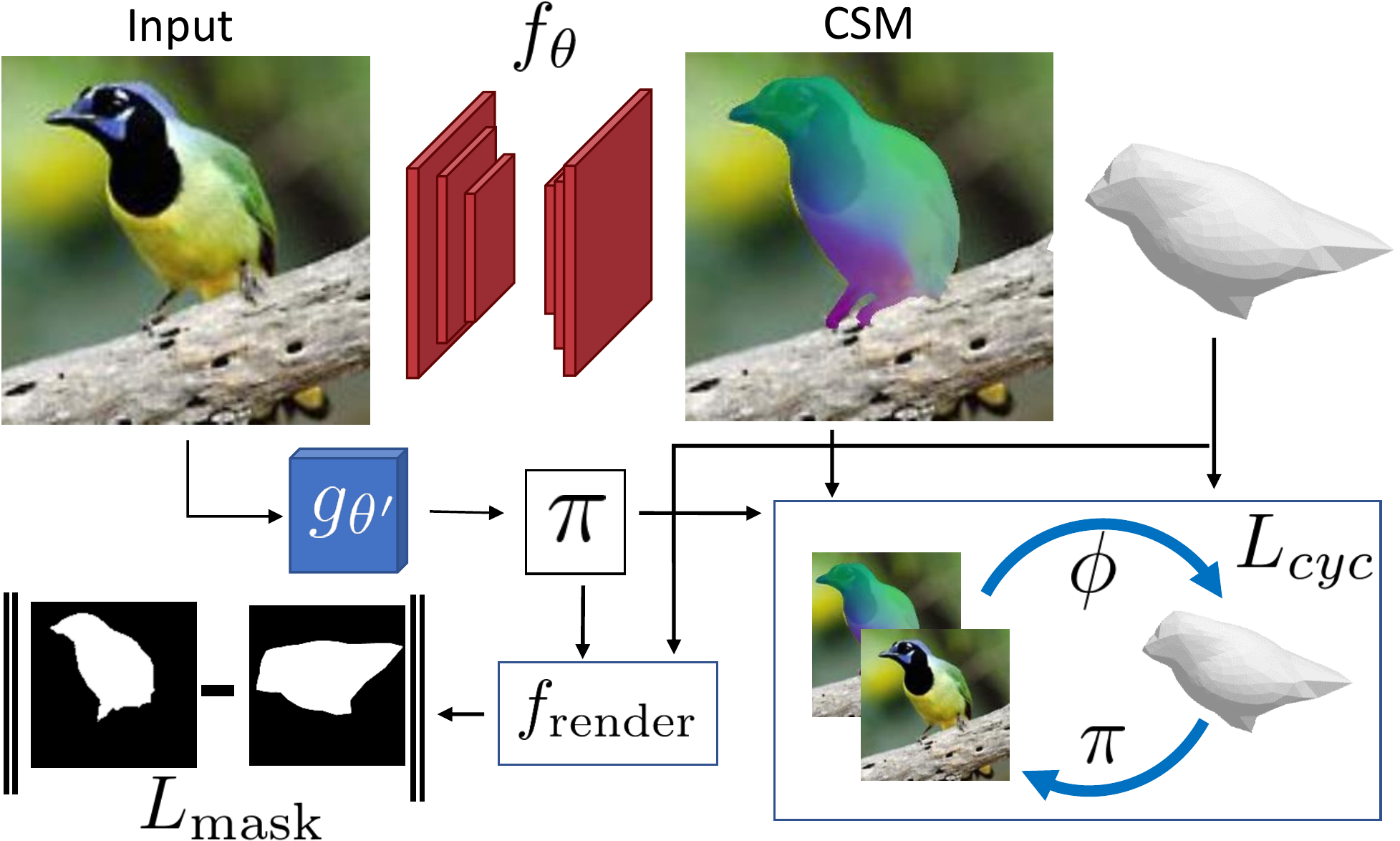}
  \caption{{\bf Overview of Training Procedure.} We train a network to predict, for each pixel on the foreground, its mapping to the canonical shape. We also jointly learn to predict camera pose, and the geometric cycle-consistency loss $L_{cyc}$ along with foreground supervision,  provides learning signal to train our system.
  }
  \label{fig:training}
\end{figure}

\subsection{Learning without Pose Supervision}
\label{subsec:relax-cam}
We have presented our approach to learn a canonical surface mapping predictor $f_{\theta}$ assuming known cameras $\pi$ for each training image. We note that our training objective is also differentiable w.r.t. the camera parameters, and we can therefore simply use predicted cameras instead of known cameras, and jointly learn pose and CSM prediction. This joint training can allow us to bypass the requirement of even camera supervision, and learn CSM prediction using only foreground mask annotations and a given template shape.

We therefore learn an additional camera-prediction CNN $g_{\theta'}$, and use the predicted cameras to learn the CSM predictor via the geometric consistency training objectives. However, to overcome certain trivial solutions, we also add a mask reprojection error, and following ~\cite{mvcTulsiani18,insafutdinov2018unsupervised} use a multi-hypothesis camera predictor to avoid local minima. Our overall training setup is depicted in Figure \ref{fig:training}.


\vspace{2mm}
\noindent {\bf Mask Re-projection Loss}.
If the only learning objective comprises of the self-consistency between camera predictions and the predicted CSMs, the networks can learn some trivial solutions \eg always predict a `frontal' camera and corresponding CSM. To avoid this we enforce that the the template shape, when viewed under a predicted camera $\pi$, should approximately match the known foreground image $I_{fg}$. To implement this loss, we use (NMR)~\cite{kato2018neural} to obtain a differentiable render $f_{render}$, that given the template shape $S$ and a camera $\pi$, renders a mask. While the poses may still be ambiguous \eg front and back facing cars, this additional mask reprojection loss allows us to circumvent the mentioned trivial solutions. This reprojection loss is defined as follows:
\begin{align}
L_{\text{mask}} = \norm{f_{\text{render}}(S, \pi) - I_{f}}^2
\end{align}

\noindent {\bf Multi-Hypothesis Pose Prediction}.
Instead of predicting a single camera  $\pi \equiv g_{\theta'}(I)$, we follow previous methods~\cite{mvcTulsiani18,insafutdinov2018unsupervised} and predict multiple hypotheses to overcome local minima. Our pose predictor outputs $\{(\pi_i, c_i)\}  \equiv g_{\theta'}(I)$ - a set of $N_{c}=8$ pose hypotheses $\pi_i$, each with an associated probability $c_i$.
We initialize the camera predictor $g_{\theta'}$ using a pre-trained ResNet-18 network~\cite{he2016deep}.



\vspace{2mm}
\noindent {\bf Overall Training Objective.}
As our pose predictor yields multiple pose hypotheses $\pi_i$, each with an associated probability $c_i$, we can train our networks by minimizing the expected loss. We denote by $L_{cyc}^{i}, L_{vis}^{i}, L_{mask}^i$ the corresponding losses under the camera prediction $\pi_i$. In addition to minimizing the expected loss over these terms, we also use an additional diversity prior $L_{div}$ to encourage diverse pose hypotheses (see appendix for details). The overall training objective using these, is:
\begin{align}
    L_{tot} &= L_{div}(g_{\theta'}(I)) + \sum_{i=1}^{N_{c}} c_i (L_{cyc}^{i} + L_{vis}^{i} +  L_{mask}^i)
\end{align}
This framework allows us to learn the canonical surface mapping function $f_{\theta}$ via geometric cycle consistency, using only foreground mask annotations in addition to the given template shape. Once the network $f_{\theta}$ is learned, we can infer a canonical surface map from any unannotated image.

\subsection{Dense Correspondences via CSM}
\label{subsec:correspondence}
We described an approach for predicting canonical surface mappings without relying on pose or keypoint annotations. This allows us to infer dense semantic correspondences given two images of the same semantic object category, because if pixels across images correspond, they should get mapped to the same region on the canonical surface. Given a (source, target) image pair $(I_s, I_t)$, let us denote by $(C_s, C_t, I_{fg}^s, I_{fg}^t)$ the corresponding predicted canonical surface mappings and foreground masks. Given these predictions, for any pixel $\mathbf{p}_s$ on $I_s$, we can infer its corresponding pixel $T_{s \rightarrow t}[\mathbf{p}_s]$ on $I_t$ by searching for the (foreground) pixel that maps closest to $\phi(C_s[\mathbf{p}_s])$.
\begin{align}
\label{eqn:correspond}
    T_{s \rightarrow t}[\mathbf{p}_s] = \text{arg}\min_{\mathbf{p}_t \in I_{fg}^t} \norm{\phi(C_s[\mathbf{p}_s]) - \phi(C_t[\mathbf{p}_t])}
\end{align}
\vspace{-2pt} 
Not only does our approach allow us to predict correspondences for pixels between two images,  it \emph{also allows us to infer regions of non-correspondence} \ie pixels in source image for which correspondences in the target image do not exist (\eg most pixels between a left and right facing bird do not correspond). We can infer these by simply denoting pixels for which the minimum distance in Eq. \ref{eqn:correspond} is above a certain threshold as not having a correspondence in the target image. This ability to infer non-correspondence is particularly challenging for self-supervised methods that generate data via random warping~\cite{kanazawa2016warpnet,thewlis2017unsupervised,Rocco_2017_CVPR} as the training pairs for these never have non-corresponding regions.


\vspace{-1mm}
\section{Experiments}

\begin{figure*}[h]
\centering
    \centering
\includegraphics[width=\textwidth]{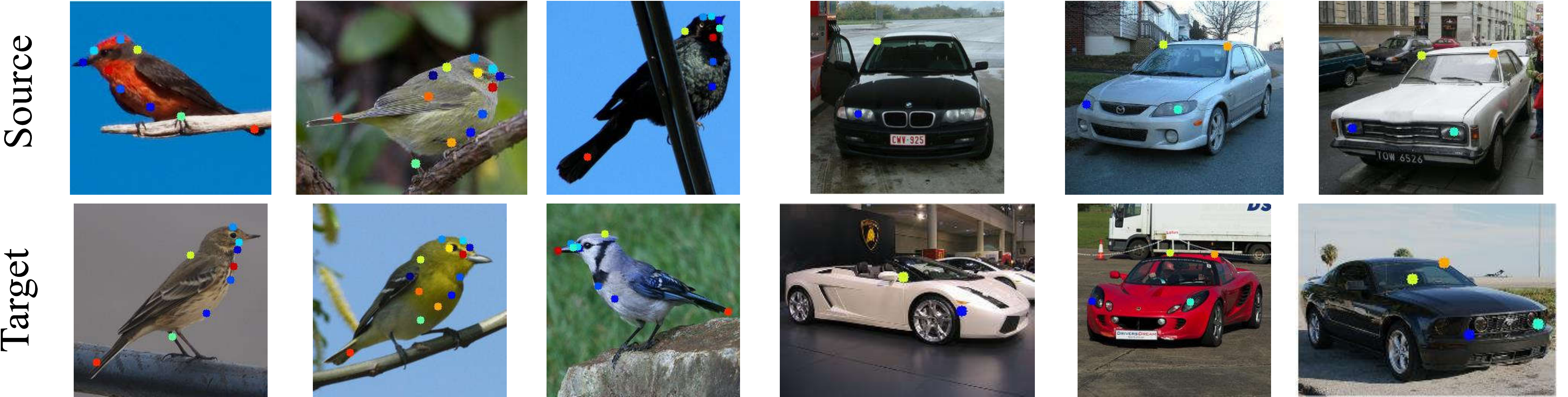}
 \captionof{figure}{{\bf Keypoint transfer results.} We show the quality of dense correspondence results by  transferring ground-truth keypoints from {\bf source} images in the top-row to {\bf target} images in the bottom-row. It is interesting to note that method is able to transfer keypoints despite significant changes in the viewpoint.} 
  \label{fig:kp_transfer}
\end{figure*}

\begin{table*}[h!]
\setlength{\tabcolsep}{0.05em}
\centering
  \scalebox{0.85}{
\begin{tabular}{ccccccc}
\addpic{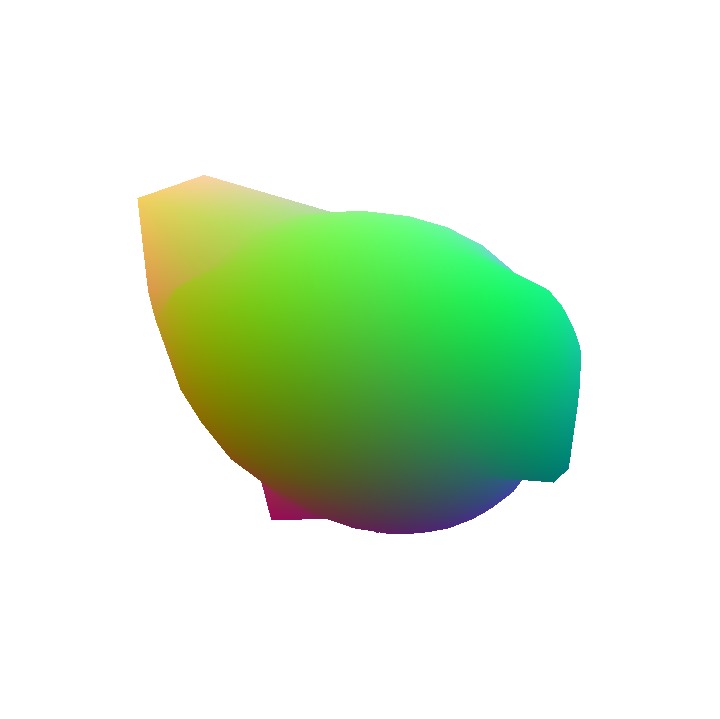}  & 
\addpic{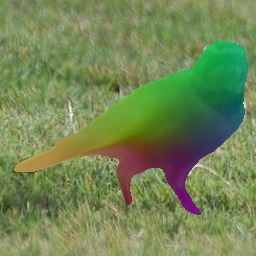}  & 
\addpic{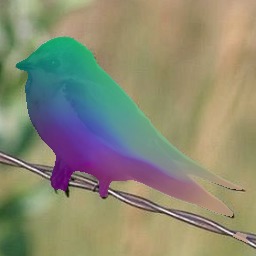} & 
\addpic{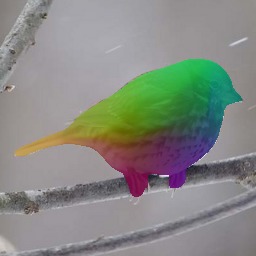}  & 
\addpic{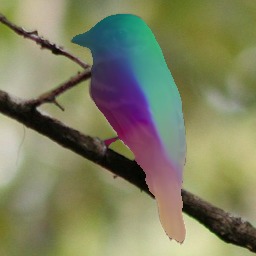}  & 
\addpic{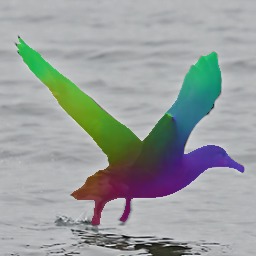} &
\addpic{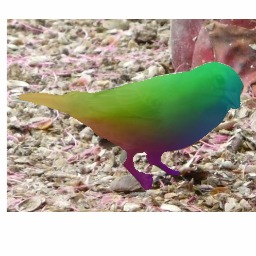} \\
\midrule
\addpic{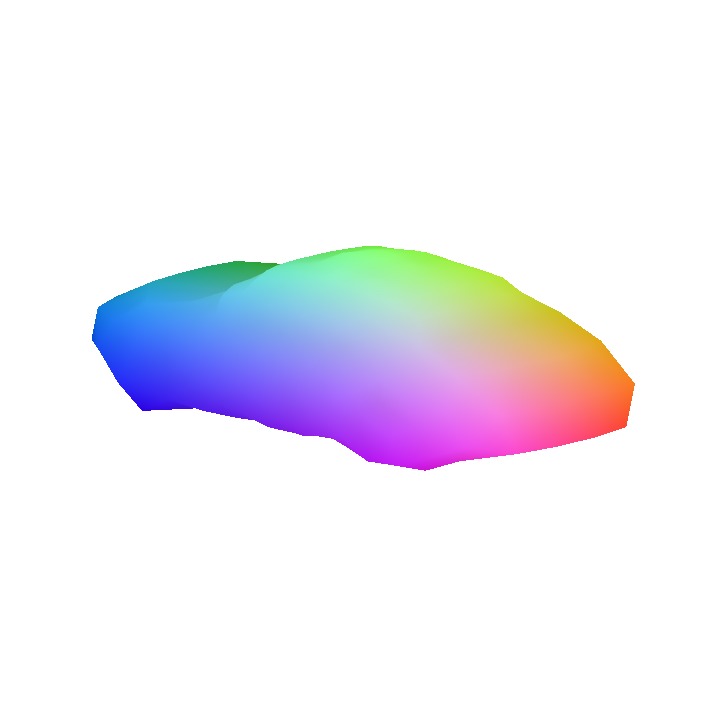}  & 
\addpic{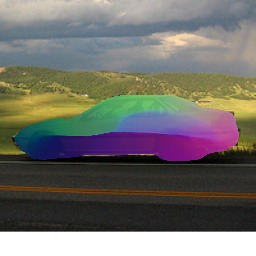}  & 
\addpic{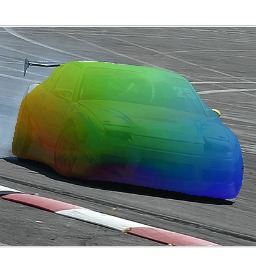}  & 
\addpic{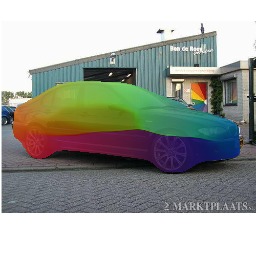}  & 
\addpic{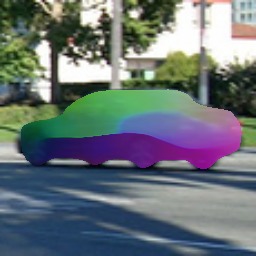}   & 
\addpic{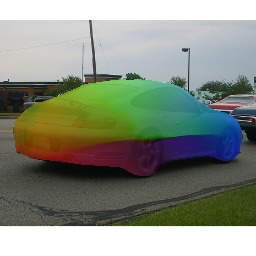}  &
\addpic{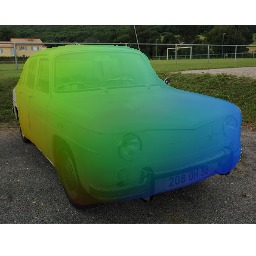} \\
\midrule
\addpic{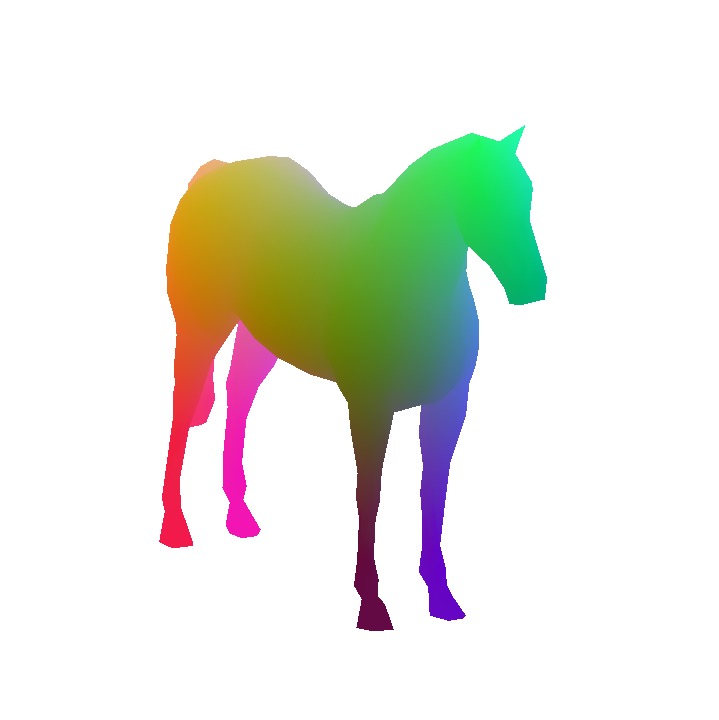}  & 
\addpic{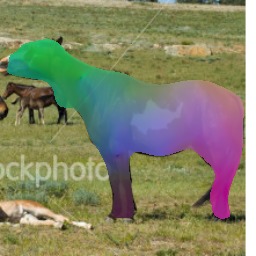}  & 
\addpic{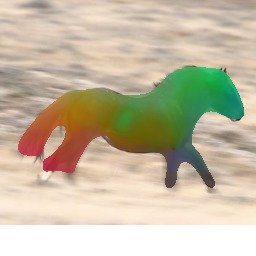}  & 
\addpic{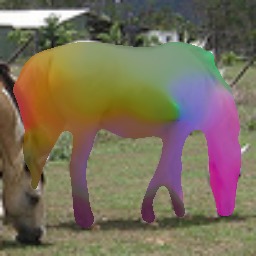}  & 
\addpic{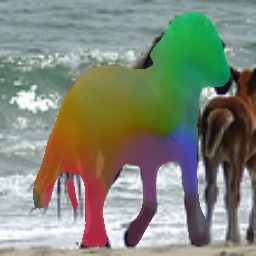}  & 
\addpic{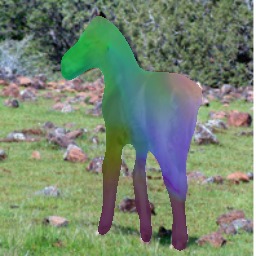}  &
\addpic{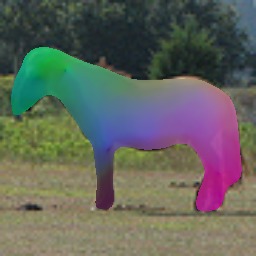} \\ 
\midrule
\addpic{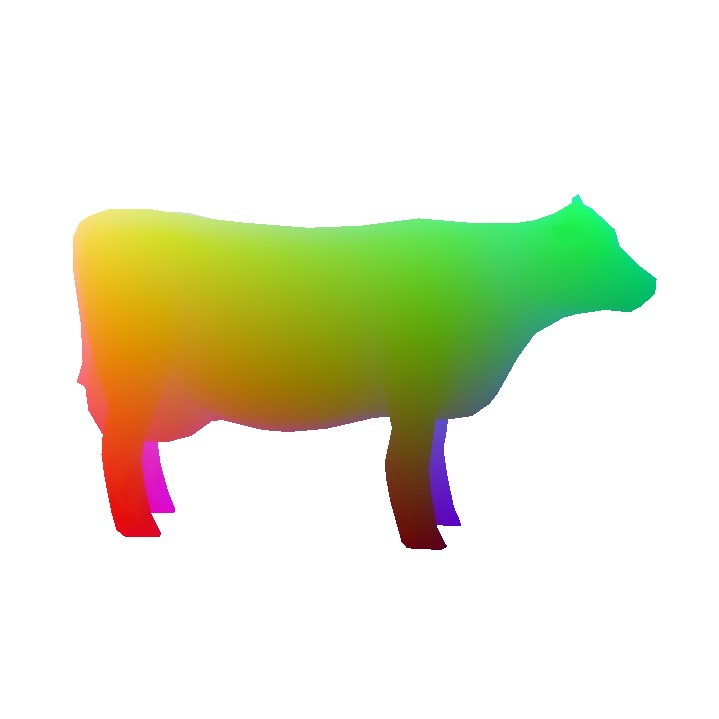}  & 
 \addpic{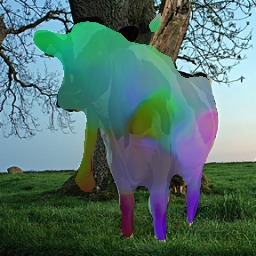}  & 
\addpic{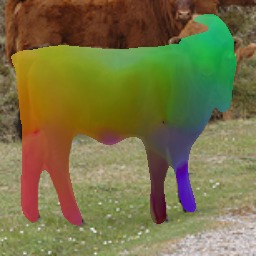}  & 
\addpic{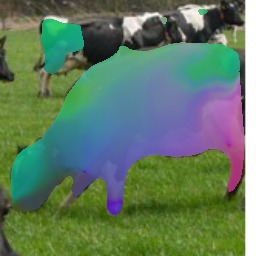}  & 
\addpic{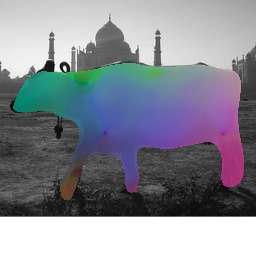}  & 
\addpic{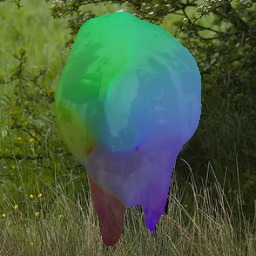} & 
\addpic{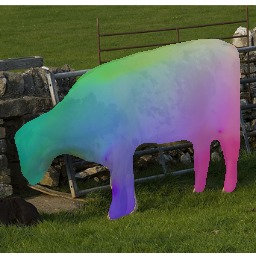} \\
\midrule
\addpic{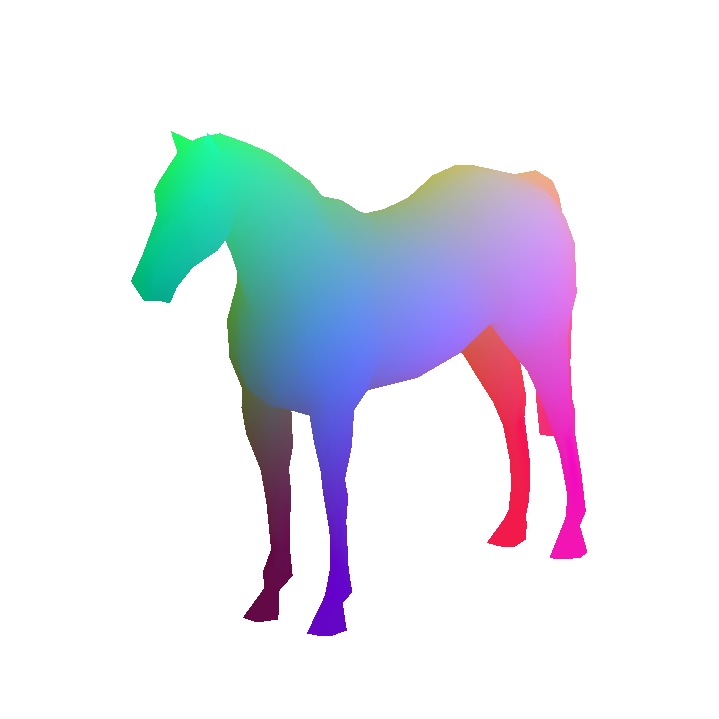} &
\addpic{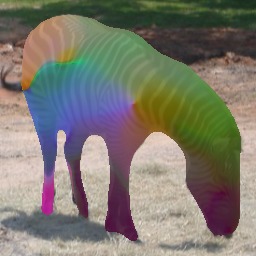}   & 
\addpic{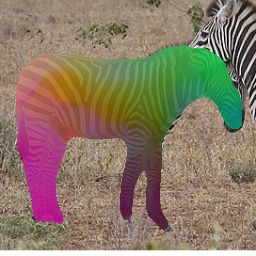}  & 
\addpic{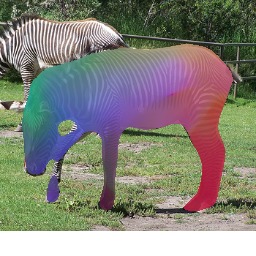} & 
\addpic{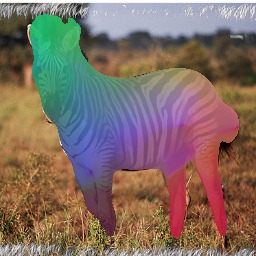}  &
\addpic{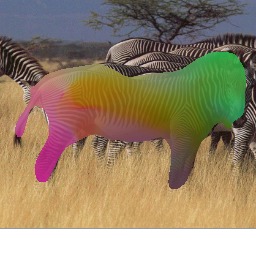}  & 
\addpic{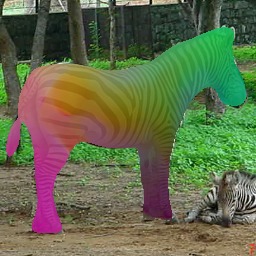}  \\
\midrule
\addpic{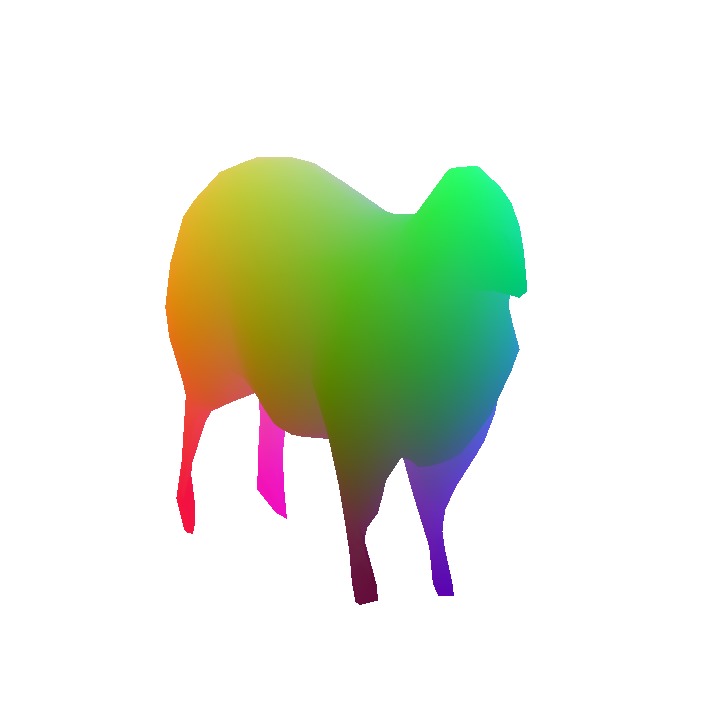} & 
\addpic{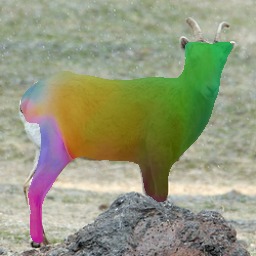}  & 
\addpic{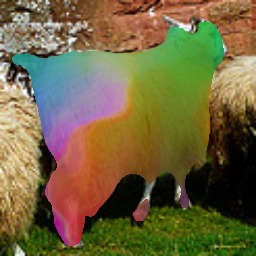}  & 
\addpic{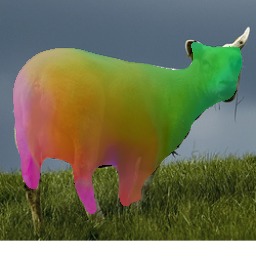}  & 
\addpic{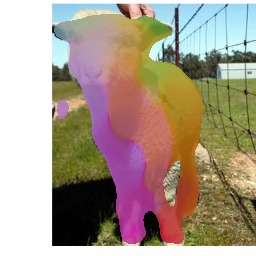}  & 
\addpic{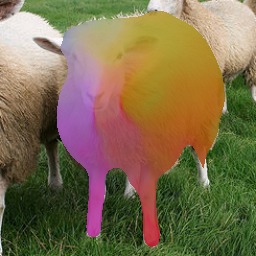}  &
\addpic{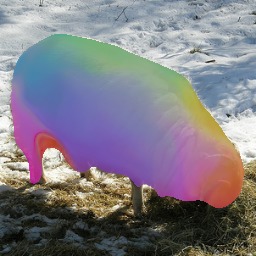} \\ 
\end{tabular}
}
\vspace{-0.1in}
\captionof{figure}{{\bf Predicted Canonical Surface mapping for six different categories}. The color at each image pixel depicts the color at the corresponding surface point on the 3D template shape in the left row. While the predictions are mostly accurate, some error modes include: a) inferring globally incorrect CSM due to pose ambiguity (\eg third horse), or b) incorrect local predictions due to missing segmentation (\eg the second sheep).}
\label{fig:qual-results-detection}
\end{table*} 

Our approach allows us to predict canonical surface mappings across generic categories. However, due to lack of annotation for the task, which is in fact our motivation for learning without supervision, it is difficult to directly evaluate the predictions. Instead, as our approach also allows us to recover correspondences across any two images (Section \ref{subsec:correspondence}), we can evaluate these using the task of keypoint transfer. This is a well-studied task by approaches that learn semantic correspondence, and we report comparisons to baselines that leverage varying degree of supervision while training. We first report these comparisons in Section \ref{subsec:eval_via_kp_tfs}, and then present results for additional generic categories (e.g. horses, sheep, cows) in Section \ref{sec:imgnetresults}, using Imagenet images with automatically obtained segmentation masks.



\subsection{Evaluation via Keypoint Transfer}
\label{subsec:eval_via_kp_tfs}
We use our learned CSM prediction models for the task of keypoint transfer -- given a source and target image pair, where the source image has some annotated keypoints, the goal is to predict the location of these keypoints in the target image. We first describe the datasets used to train our model, and then briefly survey the various baselines we compare to and then present the evaluation results. 


\subsubsection{Experimental Setup}
\noindent
{\bf Datasets.}
We use bird images from the CUB-200-2011\cite{wah2011caltech} and the car images from PASCAL3D+~\cite{xiang2014beyond} dataset for quantitative evaluation. CUB-200-2011 contains 6000 training and test images with 200 different species. Each bird has 14 annotated keypoints, a segmentation mask, and a bounding box. Note that we only use keypoint annotation at test time to evaluate our method on the task of dense correspondence as described earlier. We also train a model on the car category from PASCAL3D+~\cite{xiang2014beyond} which has over 6000 training and test images but evaluate only on cars from PASCAL VOC ~\cite{everingham2010pascal} with 12 keypoint annotations per instance. We downloaded a freely available mesh from \cite{free3d} to serve as a bird template shape, used an average of 10 Shapenet~\cite{shapenet2015} models to obtain a template shape for cars. 

\vspace{2mm} \noindent
{\bf Baselines.}
We report comparisons to several methods that leverage varying amount of supervision for learning:

\vspace{1mm} \noindent \emph{Category Specific Mesh Reconstruction (CMR)} \cite{cmrKanazawa18} learns to reconstruct the 3D shape and predict pose for a given instance, but relies on training time supervision of known keypoint locations and segmentation masks. Since a common morphable model is used across a category, we can compute the implied surface mappings via computing for each pixel, the coordinate of the mean shape that is rendered at its location (or nearest location in case of imperfect projection). We can then infer correspondences as in Section \ref{subsec:correspondence}.

\vspace{1mm} \noindent \emph{Zhou \etal} \cite{zhou2016learning} exploit a large collection of 3D synthetic models to learn dense correspondence via cycle-consistency. During training, they crucially rely on pose supervision (from PASCAL 3D+), as each cycle consists of synthetic images rendered from the same view as the real image pair. Their method outputs dense correspondences in the form of a per-pixel flow, and infers non-correspondence using a `matchability' score.

 
\vspace{1mm} \noindent {\emph{Dense Equivariance} (DE)} \cite{thewlis2017unsupervised} is a self-supervised method to learn correspondences, and does not require any pose or keypoint annotations. We re-implement this baseline such that it can exploit the annotations for object masks (see appendix for details). DE learns a per-pixel feature vector, and enforces corresponding pixels to have a similar feature. The supervision for correspondences is obtained via applying known in-plane random warps to images. During inference, we can recover the correspondence for a source pixel by searching for the most similar feature in the target image.
 
\vspace{1mm} \noindent {\emph{VGG Transfer}.} Inspired by Long \etal's~\cite{long2014convnets} observation that  generic learned features allow recovering correspondences, we designed a baseline which infers correspondence via nearest neighbours in this feature space. Specifically for a pixel in the source image we lookup its VGG feature from the conv4 layer and finds its corresponding nearest neighbour in the target image (we found these features to perform better than AlexNet used by Long \etal~\cite{long2014convnets}).
\begin{table}[h]
\setlength{\tabcolsep}{5pt} 
\centering 
\resizebox{\ifdim\width>\columnwidth
        \columnwidth
      \else
        \width
      \fi}{!}{
  \begin{tabular}{l l  l  l l l}
    \toprule
    \toprule
    Annotation & Method & \multicolumn{2}{c}{Birds} & \multicolumn{2}{c}{Cars}\\

     &  & PCK & APK & PCK & APK\\
    \midrule
    KP + Seg. Mask & CMR \cite{cmrKanazawa18} & 47.3 & 22.4 & 44.1 & 16.9    \\
    \midrule
    Pose  + Syn. Data & Zhou et. al \cite{zhou2016learning} & - & - & 37.1 & 10.5    \\ 
    Pose + Seg. Mask & CSM (ours) w/ pose & 56.0 & 30.6 & 51.2 & 21.0 \\
    \midrule
     \multirow{3}{*}{Seg. Mask}&  Dense Equi \cite{thewlis2017unsupervised}  & 34.8 & 11.1 & 31.5 & 5.7   \\
    & VGG Transfer  & 17.2 & 2.6   & 11.3 & 0.6  \\
    & CSM (ours)            & 48.0 & 22.4  & 40.0 & 11.0  \\
    \bottomrule \bottomrule
\end{tabular}
}
\vspace{-0.1in}
\caption{{\bf PCK and APK}. Percentage of correct keypoints (PCK) and Keypoint Transfer AP (APK) at $\alpha=0.1$.  See Section \ref{subsec:eval-metrics} for metric descriptions. All evaluations are on 10000 image pairs per category. Higher is better. 
}
\label{tab:pck}
\end{table}
  
\begin{table}[t]
\setlength{\tabcolsep}{0.2em} 
\centering 
  \begin{tabular}{@{}cc@{}}
  \includegraphics[width=.24\textwidth]{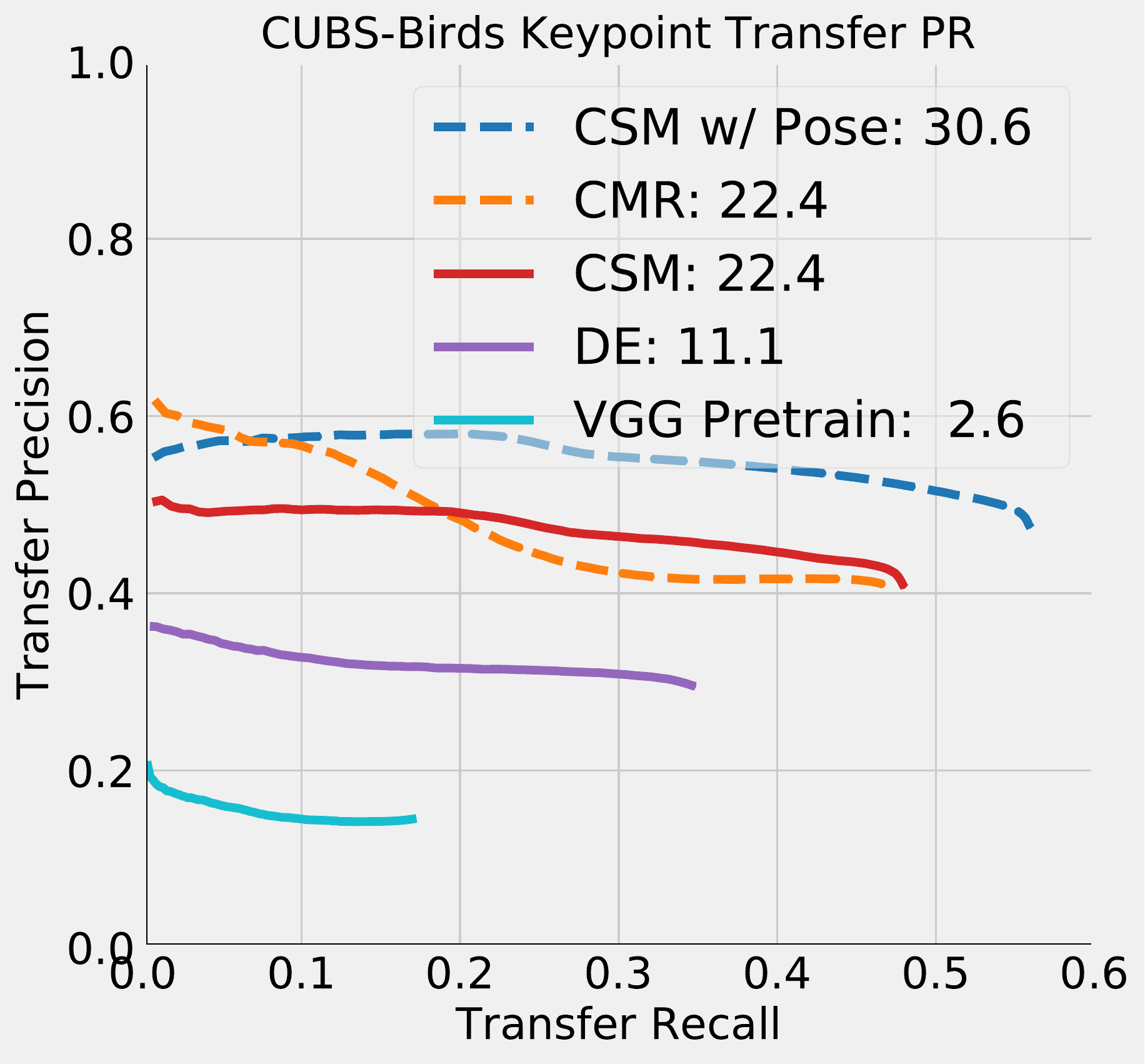} & \includegraphics[width=.24\textwidth]{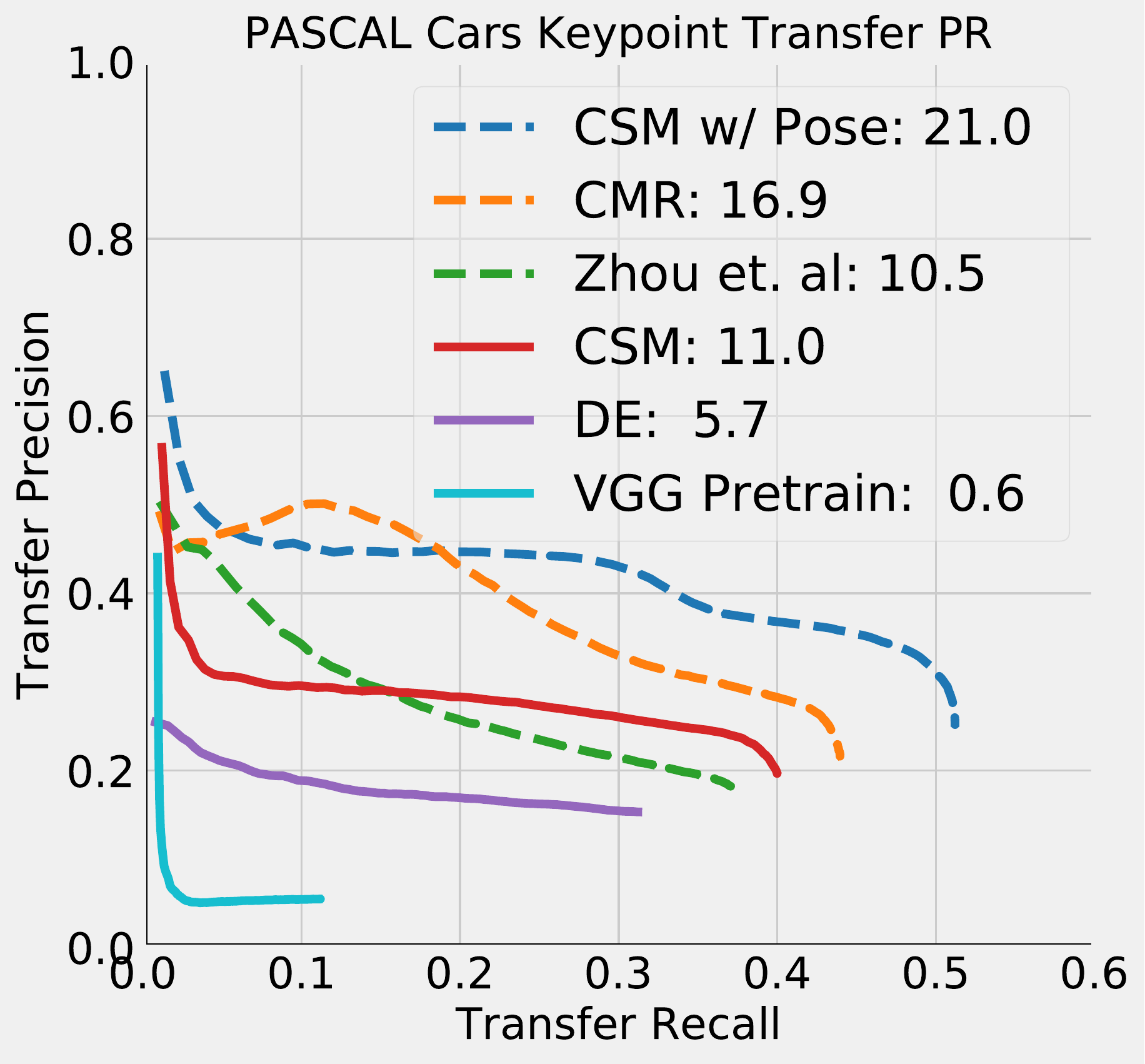}
  \end{tabular}
  \vspace{-0.1in}
  \captionof{figure}{{\bf Keypoint Transfer PR Curves.} We report the transfer precision vs recall curves for all the methods on the task of keypoint transfer. Dashed lines represent methods with pose or keypoint supervision. Solid lines denote approaches without such supervision. The area under the curve is reported in the legend for each of the plots (higher is better). The plot on left is for CUBS-Birds \cite{wah2011caltech}, and the one on the right is on cars and keypoints from PascalVOC \cite{everingham2010pascal}. See Section \ref{subsec:eval-metrics} for metric descriptions.}
\label{fig:pr-plots}
\end{table}
\vspace{-1mm}

\vspace{-1mm}
\subsubsection{Evaluation Metrics}
\label{subsec:eval-metrics}
We evaluate the various methods on two metrics: a) Percentage of Correct Keypoints (PCK), and b) Keypoint Transfer AP (APK). We use two separate metrics, because while the PCK metric evaluates the accuracy of keypoint transfer for keypoints that are visible in both, source and target image, it does not disambiguate if an approach can infer that a particular source keypoint does not correspond to any pixel on the target.
So therefore, while PCK lets us evaluate correspondence accuracy, the APK metric also lets us measure accuracy at inferring $\emph{non-correspondence}$.


\vspace{2mm} \noindent
{\bf Percentage of Correct Keypoints (PCK)}: Given a (source, target) image pair with keypoint annotations on the source, each method predicts a single estimate for the corresponding location in the target image. The PCK metric reports the mean accuracy of keypoint predictions across keypoints that are common across pairs.
A prediction is considered correct only when the predicted location lies within $\alpha*\max(h,w)$ radius around the ground truth annotation for the transfer. We report results with $\alpha=0.1$, and $h,w$ refer to height and width of the image to which the keypoints were transferred.

\vspace{2mm}\noindent
\label{subsec:apk}
{\bf Keypoint Transfer AP (APK)}: In addition to predicting a location in target image for each each keypoint in source image, this metric requires a confidence in the estimate. Ideally, if a source keypoint does not correspond in a target image, the corresponding predicted confidence should be low, whereas it should be high in case of a keypoint visible in both. Our approach and CMR~\cite{cmrKanazawa18} can rely on the (inverse) distance on the template/mean shape as a confidence measure. Zhou \etal~\cite{zhou2016learning} produce a `matchability' score, and the feature based methods `DE'~\cite{thewlis2017unsupervised} and `VGG transfer'~\cite{long2014convnets} can leverage feature similarity as confidence.
\vspace{2mm}\noindent
Given these predictions, we vary the confidence thresholds, and plot `Transfer Precision' vs `Transfer Recall' and report the area under the curve as the AP. `Transfer Recall' measures the fraction of correspondences in the ground-truth that have been recovered above the threshold (at the lowest confidence threshold, this value is similar to PCK). `Transfer Precision' measures the fraction of correspondences above the threshold that are correct (a prediction for a non-corresponding keypoint is always deemed incorrect). For a  high precision, a method should predict low confidence scores for non-corresponding keypoints. We explain these metrics in more detail in the appendix.
\vspace{-2mm}
\subsubsection{Results}
In addition to reporting the performance of our method, without any pose supervision, we also evaluate our approach when using pose supervision (denoted as `CSM w/Pose') to better compare to baselines that use similar~\cite{zhou2016learning} or more~\cite{cmrKanazawa18} annotations. However, note that all results visualization in the paper are in a setting \emph{without} known pose. We report the PCK and APK results in Table \ref{tab:pck}, and observe that our approach performs better than the alternatives. We also show the Transfer AP plots in Figure \ref{fig:pr-plots}, and note large the relative performance boost (in particular over the self-supervised method~\cite{thewlis2017unsupervised}), indicating that our approach, in addition to inferring correspondences when they exist, can realize when regions do not correspond. We also visualize some qualitative results for keypoint transfer in Figure \ref{fig:kp_transfer}.

\vspace{-1mm}
\subsection{Learning from Unannotated Image Collections}
\label{sec:imgnetresults}
\noindent
As our method does not require keypoint supervision during training, we can apply it to learn canonical surface mappings for generic classes using just category-level image collections (with automatically obtained segmentation). We use images for various categories from ImageNet~\cite{deng2009imagenet}, obtain instance segmentation using an off-the-shelf system~\cite{he2017mask}, and manually filter out instances with heavy occlusion. This results in about 1000 instances per category, and we train our CSM predictors using a per-category template model downloaded from the web (in fact, for zebras we use a horse model). We show qualitative results (on held-out images) in Figure \ref{fig:qual-results-detection} and observe that we  learn accurate mappings that also respect correspondence across instance. Please see supplementary for additional visualizations.

\vspace{-1mm}
\section{Discussion}

\noindent
We present an approach to learn canonical surface mappings for generic categories using a geometric cycle consistency objective. Our approach allows us to do so without keypoint or pose supervision, and learn CSM prediction and infer dense correspondence while only relying on foreground masks as supervision. While this is an encouraging step towards understanding the underlying 3D structure and associations across images, several challenges still remain. In particular, as we seek to explain the per-pixel predictions via a reprojection of a single rigid template, our approach is not directly applicable to categories where the shapes across instances differ significantly or undergo large articulation. It would be interesting to extend our method to also allow for predicting the underlying deformation and articulation in addition to camera transforms. Additionally, while our approach allowed relaxing correspondence supervision, it would be desirable to take a step further, and learn from unannotated image collections without foreground mask supervision. Lastly, our approach leveraged geometric cycle consistency, and videos may provide an additional learning signal by enforcing consistency of predictions through time~\cite{CVPR2019_CycleTime}.

\vspace{2mm}
\noindent \textbf{Acknowledgements.} We would like the thank the members of the CMU Visual and Robot Learning group and the anonymous reviewers for helpful discussions and feedback. ST would also like to thank Alyosha Efros for his constant nudges to `work on cycles!'. This work was partly supported by ONR MURI N000141612007 and Young Investigator Award to AG. 

{\small
\bibliographystyle{ieee_fullname}
\bibliography{references}

\begin{thebibliography}{10}\itemsep=-1pt

\bibitem{free3d}
Free3d.com.
\newblock \url{http://www.free3d.com}.

\bibitem{alp2018densepose}
R{\i}za Alp~G{\"u}ler, Natalia Neverova, and Iasonas Kokkinos.
\newblock Densepose: Dense human pose estimation in the wild.
\newblock In {\em CVPR}, 2018.

\bibitem{Anguelov:2005}
D. Anguelov, P. Srinivasan, D. Koller, S. Thrun, J. Rodgers, and J. Davis.
\newblock {SCAPE}: {S}hape {C}ompletion and {A}nimation of {PE}ople.
\newblock {\em SIGGRAPH}, 2005.

\bibitem{BlanzVetter}
Volker Blanz and Thomas Vetter.
\newblock A morphable model for the synthesis of 3d faces.
\newblock In {\em SIGGRAPH}, 1999.

\bibitem{cashman2013shape}
Thomas~J Cashman and Andrew~W Fitzgibbon.
\newblock What shape are dolphins? building 3d morphable models from 2d images.
\newblock {\em TPAMI}, 2013.

\bibitem{shapenet2015}
Angel~X. Chang, Thomas Funkhouser, Leonidas Guibas, Pat Hanrahan, Qixing Huang,
  Zimo Li, Silvio Savarese, Manolis Savva, Shuran Song, Hao Su, Jianxiong Xiao,
  Li Yi, and Fisher Yu.
\newblock {ShapeNet: An Information-Rich 3D Model Repository}.
\newblock Technical Report arXiv:1512.03012 [cs.GR], Stanford University ---
  Princeton University --- Toyota Technological Institute at Chicago, 2015.

\bibitem{choy2016universal}
Christopher~B Choy, JunYoung Gwak, Silvio Savarese, and Manmohan Chandraker.
\newblock Universal correspondence network.
\newblock In {\em NeurIPS}, 2016.

\bibitem{deng2009imagenet}
Jia Deng, Wei Dong, Richard Socher, Li-Jia Li, Kai Li, and Li Fei-Fei.
\newblock Imagenet: A large-scale hierarchical image database.
\newblock In {\em CVPR}, 2009.

\bibitem{everingham2010pascal}
Mark Everingham, Luc Van~Gool, Christopher~KI Williams, John Winn, and Andrew
  Zisserman.
\newblock The pascal visual object classes (voc) challenge.
\newblock {\em IJCV}, 2010.

\bibitem{garg2016unsupervised}
Ravi Garg, Vijay~Kumar BG, Gustavo Carneiro, and Ian Reid.
\newblock Unsupervised cnn for single view depth estimation: Geometry to the
  rescue.
\newblock In {\em ECCV}. Springer, 2016.

\bibitem{godard2017unsupervised}
Cl{\'e}ment Godard, Oisin Mac~Aodha, and Gabriel~J Brostow.
\newblock Unsupervised monocular depth estimation with left-right consistency.
\newblock In {\em CVPR}, 2017.

\bibitem{gwak2017weakly}
JunYoung Gwak, Christopher~B Choy, Animesh Garg, Manmohan Chandraker, and
  Silvio Savarese.
\newblock Weakly supervised 3d reconstruction with adversarial constraint.
\newblock In {\em 3DV}, 2017.

\bibitem{ham2016proposal}
Bumsub Ham, Minsu Cho, Cordelia Schmid, and Jean Ponce.
\newblock Proposal flow.
\newblock In {\em CVPR}, 2016.

\bibitem{he2017mask}
Kaiming He, Georgia Gkioxari, Piotr Doll{\'a}r, and Ross Girshick.
\newblock Mask r-cnn.
\newblock In {\em ICCV}, 2017.

\bibitem{he2016deep}
Kaiming He, Xiangyu Zhang, Shaoqing Ren, and Jian Sun.
\newblock Deep residual learning for image recognition.
\newblock In {\em CVPR}, 2016.

\bibitem{insafutdinov2018unsupervised}
Eldar Insafutdinov and Alexey Dosovitskiy.
\newblock Unsupervised learning of shape and pose with differentiable point
  clouds.
\newblock In {\em NeurIPS}, 2018.

\bibitem{kanazawa2016warpnet}
Angjoo Kanazawa, David~W Jacobs, and Manmohan Chandraker.
\newblock Warpnet: Weakly supervised matching for single-view reconstruction.
\newblock In {\em CVPR}, 2016.

\bibitem{cmrKanazawa18}
Angjoo Kanazawa, Shubham Tulsiani, Alexei~A. Efros, and Jitendra Malik.
\newblock Learning category-specific mesh reconstruction from image
  collections.
\newblock In {\em ECCV}, 2018.

\bibitem{CSDM}
Abhishek Kar, Shubham Tulsiani, Jo{\~{a}}o Carreira, and Jitendra Malik.
\newblock Category-specific object reconstruction from a single image.
\newblock In {\em CVPR}, 2015.

\bibitem{kato2018neural}
Hiroharu Kato, Yoshitaka Ushiku, and Tatsuya Harada.
\newblock Neural 3d mesh renderer.
\newblock In {\em CVPR}, 2018.

\bibitem{khamis2015learning}
Sameh Khamis, Jonathan Taylor, Jamie Shotton, Cem Keskin, Shahram Izadi, and
  Andrew Fitzgibbon.
\newblock Learning an efficient model of hand shape variation from depth
  images.
\newblock In {\em CVPR}, 2015.

\bibitem{kim2013deformable}
Jaechul Kim, Ce Liu, Fei Sha, and Kristen Grauman.
\newblock Deformable spatial pyramid matching for fast dense correspondences.
\newblock In {\em CVPR}, 2013.

\bibitem{kim2017learning}
Taeksoo Kim, Moonsu Cha, Hyunsoo Kim, Jung~Kwon Lee, and Jiwon Kim.
\newblock Learning to discover cross-domain relations with generative
  adversarial networks.
\newblock In {\em ICML}, 2017.

\bibitem{kingma2014adam}
Diederik~P Kingma and Jimmy Ba.
\newblock Adam: A method for stochastic optimization.
\newblock {\em arXiv preprint arXiv:1412.6980}, 2014.

\bibitem{liu2011sift}
Ce Liu, Jenny Yuen, and Antonio Torralba.
\newblock Sift flow: Dense correspondence across scenes and its applications.
\newblock {\em TPAMI}, 2011.

\bibitem{long2014convnets}
Jonathan~L Long, Ning Zhang, and Trevor Darrell.
\newblock Do convnets learn correspondence?
\newblock In {\em NeurIPS}, 2014.

\bibitem{SMPL}
Matthew Loper, Naureen Mahmood, Javier Romero, Gerard Pons-Moll, and Michael~J.
  Black.
\newblock {SMPL}: A skinned multi-person linear model.
\newblock {\em SIGGRAPH Asia}, 2015.

\bibitem{pons2015metric}
Gerard Pons-Moll, Jonathan Taylor, Jamie Shotton, Aaron Hertzmann, and Andrew
  Fitzgibbon.
\newblock Metric regression forests for correspondence estimation.
\newblock {\em IJCV}, 2015.

\bibitem{praun2003spherical}
Emil Praun and Hugues Hoppe.
\newblock Spherical parametrization and remeshing.
\newblock In {\em TOG}, 2003.

\bibitem{Narapureddy-2018-105893}
N~Dinesh Reddy, Minh Vo, and Srinivasa~G. Narasimhan.
\newblock Carfusion: Combining point tracking and part detection for dynamic 3d
  reconstruction of vehicle.
\newblock In {\em CVPR}, 2018.

\bibitem{rezende2016unsupervised}
Danilo~Jimenez Rezende, SM~Ali Eslami, Shakir Mohamed, Peter Battaglia, Max
  Jaderberg, and Nicolas Heess.
\newblock Unsupervised learning of 3d structure from images.
\newblock In {\em NeurIPS}, 2016.

\bibitem{rhodin2018unsupervised}
Helge Rhodin, Mathieu Salzmann, and Pascal Fua.
\newblock Unsupervised geometry-aware representation for 3d human pose
  estimation.
\newblock In {\em ECCV}, 2018.

\bibitem{Rocco_2017_CVPR}
Ignacio Rocco, Relja Arandjelovic, and Josef Sivic.
\newblock Convolutional neural network architecture for geometric matching.
\newblock In {\em CVPR}, 2017.

\bibitem{rocco2018end}
Ignacio Rocco, Relja Arandjelovi{\'c}, and Josef Sivic.
\newblock End-to-end weakly-supervised semantic alignment.
\newblock In {\em CVPR}, 2018.

\bibitem{ronneberger2015u}
Olaf Ronneberger, Philipp Fischer, and Thomas Brox.
\newblock U-net: Convolutional networks for biomedical image segmentation.
\newblock In {\em MICCAI}, 2015.

\bibitem{schmidt2017self}
Tanner Schmidt, Richard Newcombe, and Dieter Fox.
\newblock Self-supervised visual descriptor learning for dense correspondence.
\newblock {\em IEEE Robotics and Automation Letters}, 2017.

\bibitem{simon2017hand}
Tomas Simon, Hanbyul Joo, Iain Matthews, and Yaser Sheikh.
\newblock Hand keypoint detection in single images using multiview
  bootstrapping.
\newblock In {\em CVPR}, 2017.

\bibitem{taylor2012vitruvian}
Jonathan Taylor, Jamie Shotton, Toby Sharp, and Andrew Fitzgibbon.
\newblock The vitruvian manifold: Inferring dense correspondences for one-shot
  human pose estimation.
\newblock In {\em CVPR}, 2012.

\bibitem{taylor2014user}
Jonathan Taylor, Richard Stebbing, Varun Ramakrishna, Cem Keskin, Jamie
  Shotton, Shahram Izadi, Aaron Hertzmann, and Andrew Fitzgibbon.
\newblock User-specific hand modeling from monocular depth sequences.
\newblock In {\em CVPR}, 2014.

\bibitem{thewlis2017unsupervised}
James Thewlis, Hakan Bilen, and Andrea Vedaldi.
\newblock Unsupervised learning of object frames by dense equivariant image
  labelling.
\newblock In {\em NeurIPS}, 2017.

\bibitem{Thompson}
D'Arcy Thompson.
\newblock {\em On Growth and Form}.
\newblock Cambridge Univ. Press, 1917.

\bibitem{mvcTulsiani18}
Shubham Tulsiani, Alexei~A. Efros, and Jitendra Malik.
\newblock Multi-view consistency as supervisory signal for learning shape and
  pose prediction.
\newblock In {\em CVPR}, 2018.

\bibitem{drcTulsiani17}
Shubham Tulsiani, Tinghui Zhou, Alexei~A. Efros, and Jitendra Malik.
\newblock Multi-view supervision for single-view reconstruction via
  differentiable ray consistency.
\newblock In {\em CVPR}, 2017.

\bibitem{tung2017self}
Hsiao-Yu Tung, Hsiao-Wei Tung, Ersin Yumer, and Katerina Fragkiadaki.
\newblock Self-supervised learning of motion capture.
\newblock In {\em NeurIPS}, 2017.

\bibitem{wah2011caltech}
Catherine Wah, Steve Branson, Peter Welinder, Pietro Perona, and Serge
  Belongie.
\newblock The caltech-ucsd birds-200-2011 dataset.
\newblock 2011.

\bibitem{Wang_UnsupICCV2017}
Xiaolong Wang, Kaiming He, and Abhinav Gupta.
\newblock Transitive invariance for self-supervised visual representation
  learning.
\newblock In {\em ICCV}, 2017.

\bibitem{CVPR2019_CycleTime}
Xiaolong Wang, Allan Jabri, and Alexei~A. Efros.
\newblock Learning correspondence from the cycle-consistency of time.
\newblock In {\em CVPR}, 2019.

\bibitem{3dinterpreter}
Jiajun Wu, Tianfan Xue, Joseph~J Lim, Yuandong Tian, Joshua~B Tenenbaum,
  Antonio Torralba, and William~T Freeman.
\newblock Single image 3d interpreter network.
\newblock In {\em ECCV}, 2016.

\bibitem{xiang2014beyond}
Yu Xiang, Roozbeh Mottaghi, and Silvio Savarese.
\newblock Beyond pascal: A benchmark for 3d object detection in the wild.
\newblock In {\em WACV}, 2014.

\bibitem{yan2016perspective}
Xinchen Yan, Jimei Yang, Ersin Yumer, Yijie Guo, and Honglak Lee.
\newblock Perspective transformer nets: Learning single-view 3d object
  reconstruction without 3d supervision.
\newblock In {\em NeurIPS}, 2016.

\bibitem{zeng20173dmatch}
Andy Zeng, Shuran Song, Matthias Nie{\ss}ner, Matthew Fisher, Jianxiong Xiao,
  and Thomas Funkhouser.
\newblock 3dmatch: Learning local geometric descriptors from rgb-d
  reconstructions.
\newblock In {\em CVPR}, 2017.

\bibitem{zhou2017unsupervised}
Tinghui Zhou, Matthew Brown, Noah Snavely, and David~G Lowe.
\newblock Unsupervised learning of depth and ego-motion from video.
\newblock In {\em CVPR}, 2017.

\bibitem{zhou2015flowweb}
Tinghui Zhou, Yong Jae~Lee, Stella~X Yu, and Alyosha~A Efros.
\newblock Flowweb: Joint image set alignment by weaving consistent, pixel-wise
  correspondences.
\newblock In {\em CVPR}, 2015.

\bibitem{zhou2016learning}
Tinghui Zhou, Philipp Kr{\"a}henb{\"u}hl, Mathieu Aubry, Qixing Huang, and
  Alexei~A. Efros.
\newblock Learning dense correspondence via 3d-guided cycle consistency.
\newblock In {\em CVPR}, 2016.

\bibitem{zhu2017unpaired}
Jun-Yan Zhu, Taesung Park, Phillip Isola, and Alexei~A Efros.
\newblock Unpaired image-to-image translation using cycle-consistent
  adversarial networks.
\newblock In {\em ICCV}, 2017.

\end{thebibliography}
}

\clearpage
\renewcommand{\thesection}{\Alph{section}}
\setcounter{section}{0}
\vspace{-4mm}

\section{Training Details}
\subsection{Network Archicture}

We use a 5 Layer UNet~\cite{ronneberger2015u} with 4 $\times 4$ convolution at each layer. The UNet takes input an image and learns to predict an unit-vector which parameterizes $u,v$. Along with that we also train the UNet to predict a segmentation of the object which is necessary for keypoint evaluations. We train our networks for over 200 epochs on all the datasets independently. We use Adam~\cite{kingma2014adam} for optimization of our neural network with a learning rate of $10^{-4}$.

\subsection{Optimization}
{\bf Pose Prediction} We predict $N$ (=8) possible hypothesis for pose given an image. We initialize the poses such that they span a wide spectrum during start of the training. We add an additional loss to encourage diversity and to ensure there is no mode collapse. The diversity loss consists of two terms:
\begin{itemize}
    \item We add an entropy term over the probabilities of hypothesis $c_{i}$ which prevents mode collapse, and encourages exploration. This is equivalent to minimizing $\sum_{i}^{N} c_{i} \log(c_{i})$
    \item We maximize a pair-wise distance between predicted rotations, $\text{Dist}(r_{i}, r_{j})$ for all the predicted hypothesis of an instance. This is equivalent to minimizing $\sum_{i=1}^{N} \sum_{j=1, j\neq i}^{N} \text{Dist}(r_{i}, r_{j})$
\end{itemize}

\section{Evaluation Metrics}
{\bf Keypoint Transfer AP (APK).} Any keypoint transfer method given two images as input helps us infer how keypoints transfer from a source image to a target image. The method give two outputs for every keypoint a) transferred keypoint location b) confidence score. A keypoint transfer is successful if the confidence score of the method for the transfer is high and the error for the transfer is less than $d=\alpha \times \max(h,w)$, where $h,w$ represent height and width respectively. For any method we create several confidence thresholds compute the following metrics. Let us consider we have a lot of image-pairs where we have only $N_{\text{pair}}$ keypoint correspondences between source and target. For any given confidence threshold $t$ following are the two cases:-
\begin{enumerate}
    \item True Positive (TP): The confidence for the correspondence was above $t$, and the transfer error is less than $d$.
    \item False Positive (FP): The confidence for the correspondence was above $t$, but either the given keypoint does not exist on the target image, or our transfer error is more than $d$.
\end{enumerate}
We compute transfer precision and transfer recall as follows
\begin{align*}
    \text{Transfer Precision} &= \dfrac{N_{TP}}{N_{TP}+ N_{FP}}\\
    \text{Transfer Recall} &= \dfrac{N_{TP}}{N_{\text{pair}}}\\
\end{align*}
Here, $N_{TP}$ represents number of True Positives and $N_{FP}$ represents number of False Positives. We create the plots for transfer precision vs transfer recall as shown in the Figure {\color{red} 7} in the main manuscript. Area under such a plot represents AP and we report performance on the same in Table {\color{red} 1} in the main manuscript.

\section{Ablations}
We investigate the importance of: a) the visibility loss (-vis), b) the use of foreground pixels in $L_{\text{consistency}}$  loss  (-mask). We report our quantitative evaluations in Table \ref{tab:ablation}. We observe that visibility constraint is important, and the ablations show a  drop in average performance across both the metrics if this loss is excluded during training. Our CSM model is trained with $L_{\text{consistency}}$ loss only on foreground pixels, and the experiments denoted by ( -mask) ablate this and do not use segmentation mask while computing the losses. We observe that using cycle and visibility loss over all the pixels in the image does not significantly affect performance. Note that the mask supervision is still critical for the reprojection loss that helps resolve degenerate solutions as described earlier, and the predicted masks are also used for correspondence transfer as in Equation {\color{red} 6} in the main manuscript.

\begin{table}[h]
\setlength{\tabcolsep}{13pt} 
\centering 

\scalebox{0.8}{
  \begin{tabular}{l l  l  l l }
    \toprule
    \toprule
   Method & \multicolumn{2}{c}{Birds} & \multicolumn{2}{c}{Cars}\\

    & PCK & APK & PCK & APK\\
    \midrule
    CSM w/ pose        & 56.0 & 30.6 & 51.2 & 21.0 \\
    CSM w/ pose - vis  & 57.0 & 31.9 & 42.5 & 12.8 \\
    CSM w/ pose - mask & 53.2 & 27.4 & 51.2 & 21.5 \\
    \midrule
    CSM               & 48.0 & 22.4  & 40.0 & 11.0  \\
    CSM - vis               & 43.1 & 18.3 & 33.0  & 7.1  \\
    CSM - mask               & 45.1  & 20.0  & 40.0 & 10.9  \\
    \bottomrule
\end{tabular}
}
\caption{\textbf{Ablations.} The settings with (-vis) indicate results if visibility loss is not enforced. The settings with (-mask) refer to enforcing $L_{\text{consistency}}$ loss on all pixels, and not just foreground ones, though the reprojection loss still leveraged mask supervision.}
\label{tab:ablation}
\end{table}


\section{Results on Internet Videos}
In the supplementary video we show results of our method on several videos.
The color map on the video sequences shows correspondence to the template shape -- shown at the top right of the frame. This helps us understand and visualize intra-frame correspondences. They also show the consistency of our predictions across frames. For instance, similar colors for the tails of two birds indicates that these pixels map to similar points on the template shape.  We see few snapshots from the videos  in the Fig \ref{fig:video}. It is important to note that since we are using segmentation masks from pre-trained Mask-RCNN, the failure modes of Mask-RCNN become our failure modes. We observe that false-detections and failure to detect the instance in certain frames results in absence of CSM. Furthermore, since we only train using isolated untruncated and unoccluded objects, our predictions are often inaccurate if objects overlap or are not fully visible.

It is important to note that we do not apply any smoothing or consistency across frames. Our method operates on all the frames in the video independently.
\begin{table*}
\setlength{\tabcolsep}{0.05em}
\centering
  \scalebox{0.85}{
\begin{tabular}{cc}
\addpicBig{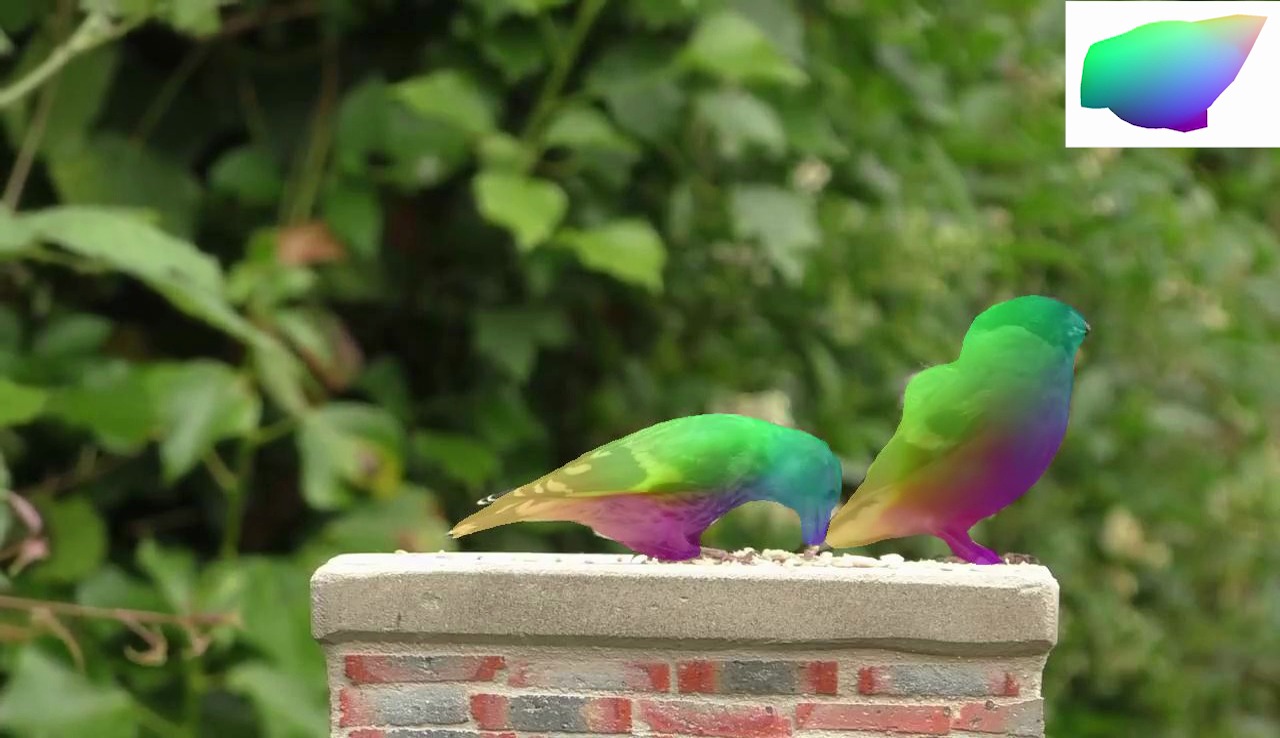} &
\addpicBig{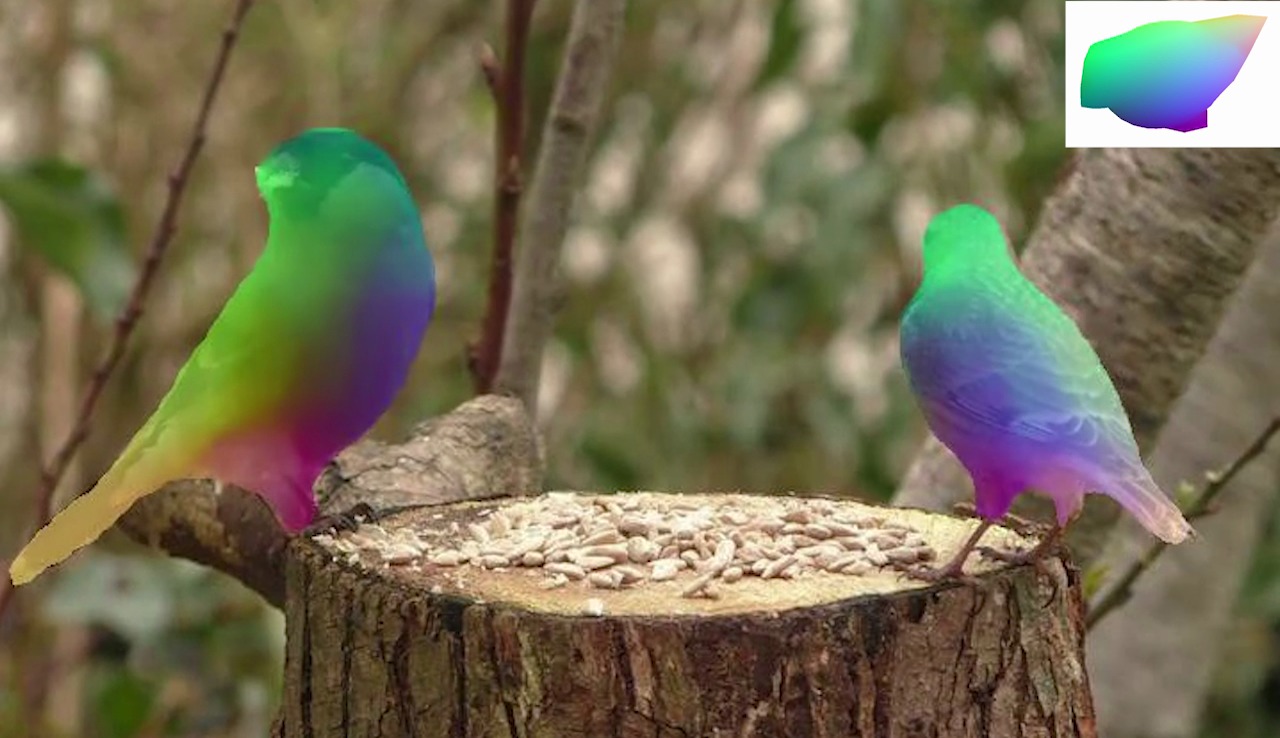} \\
\addpicBig{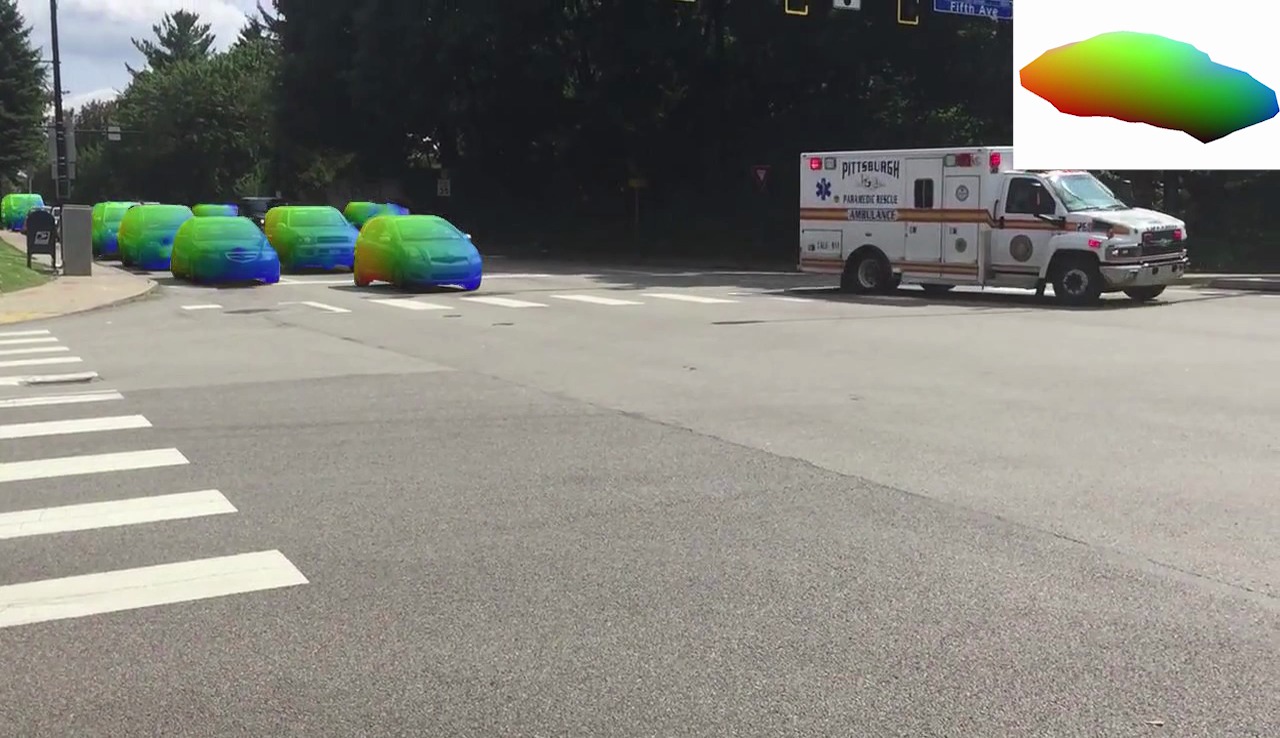} &
\addpicBig{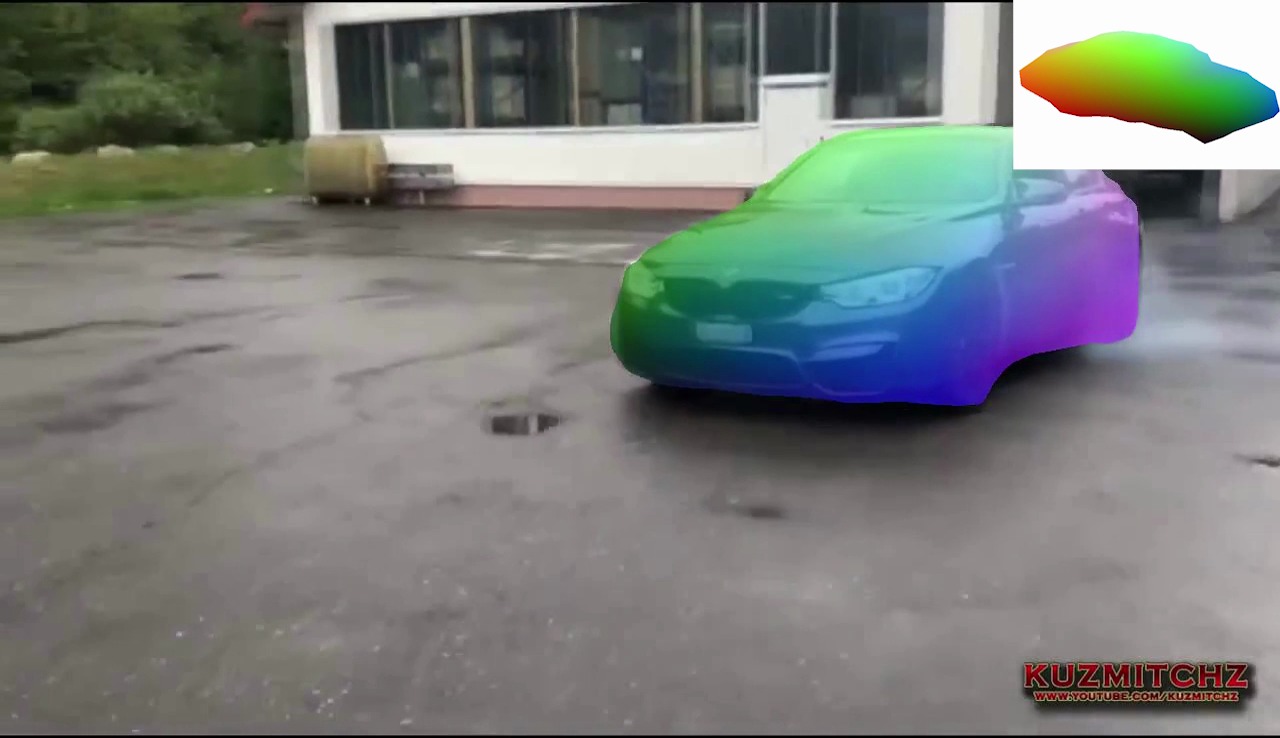} \\
\addpicBig{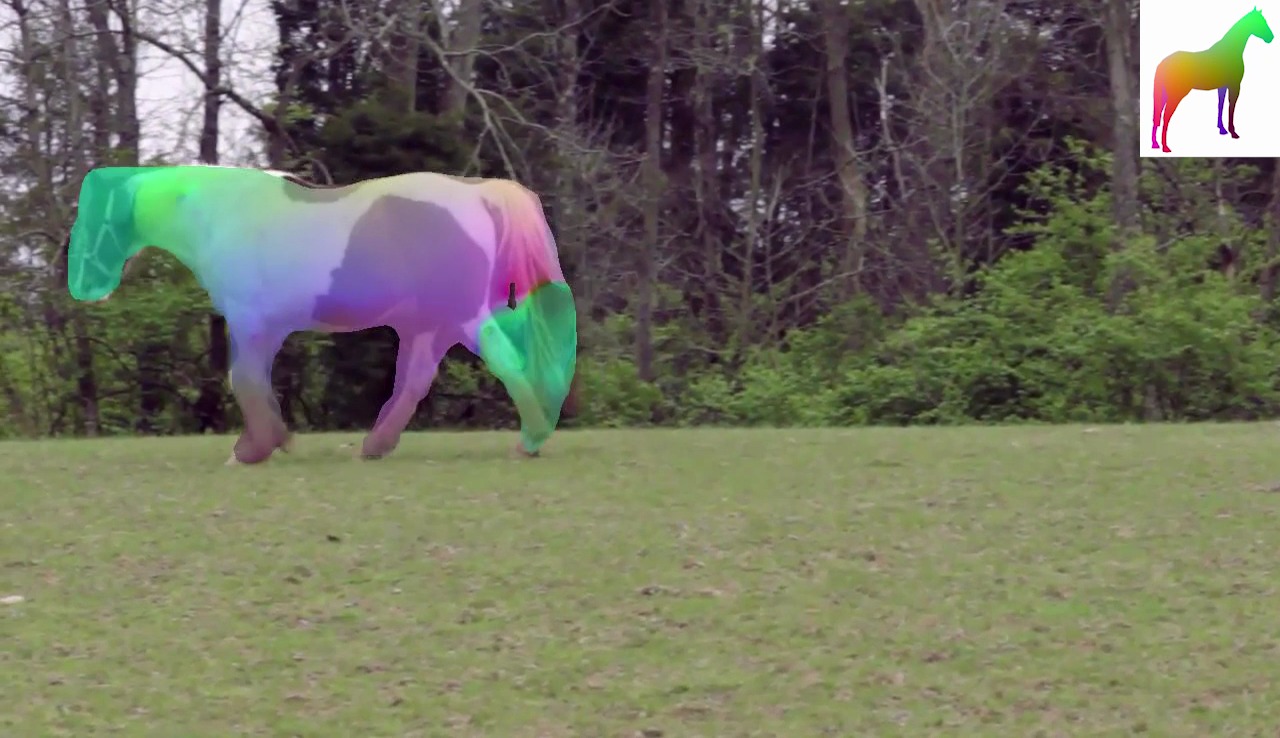} &
\addpicBig{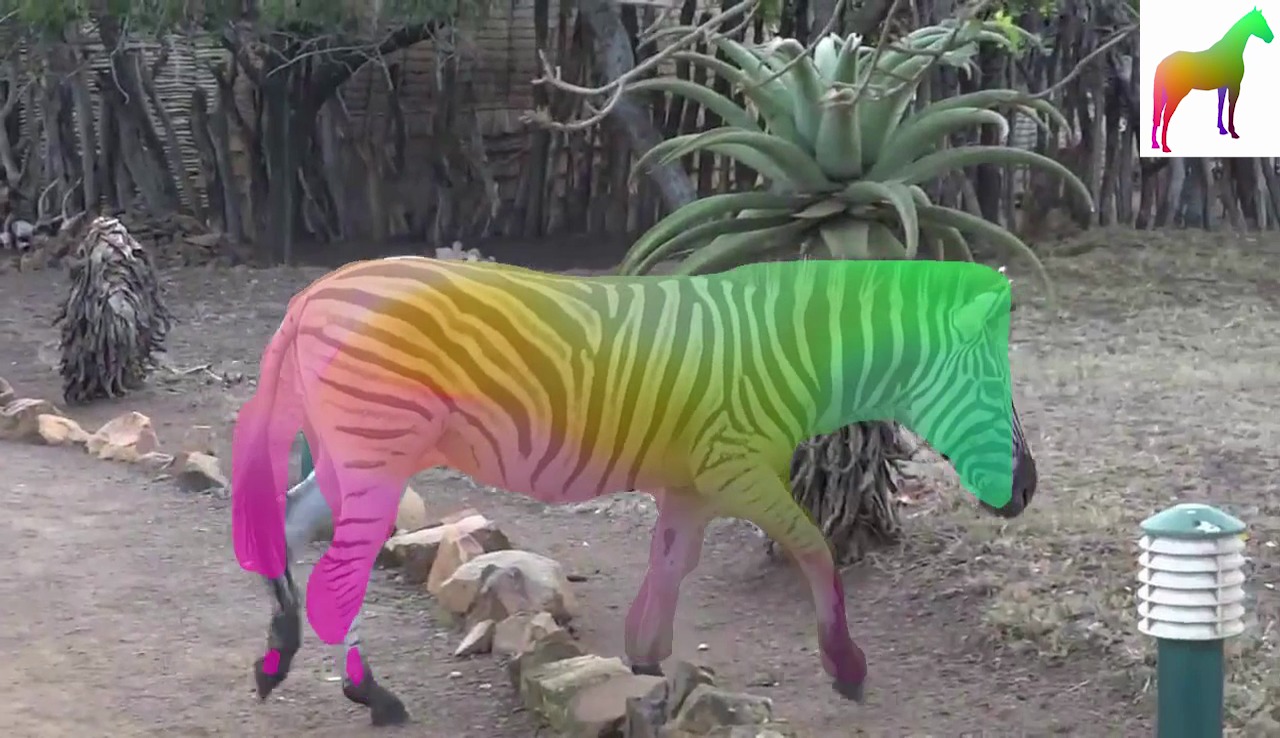} \\
\addpicBig{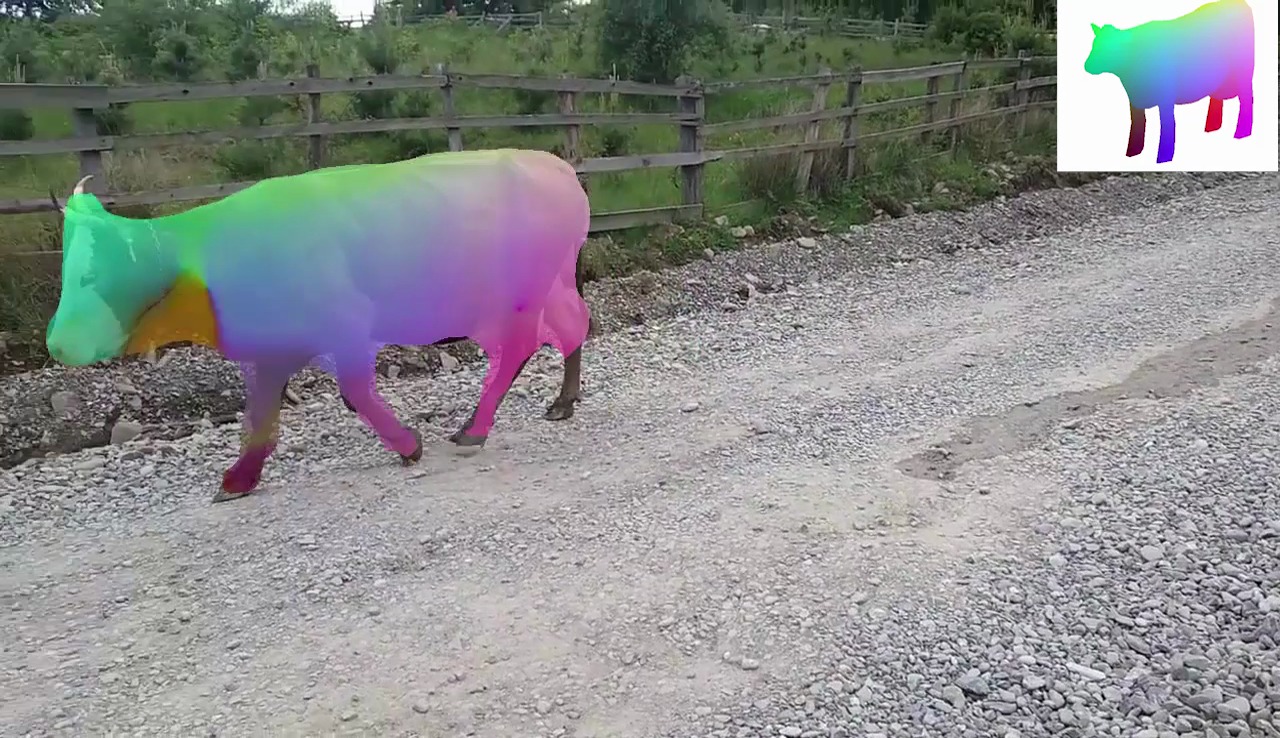} &
\addpicBig{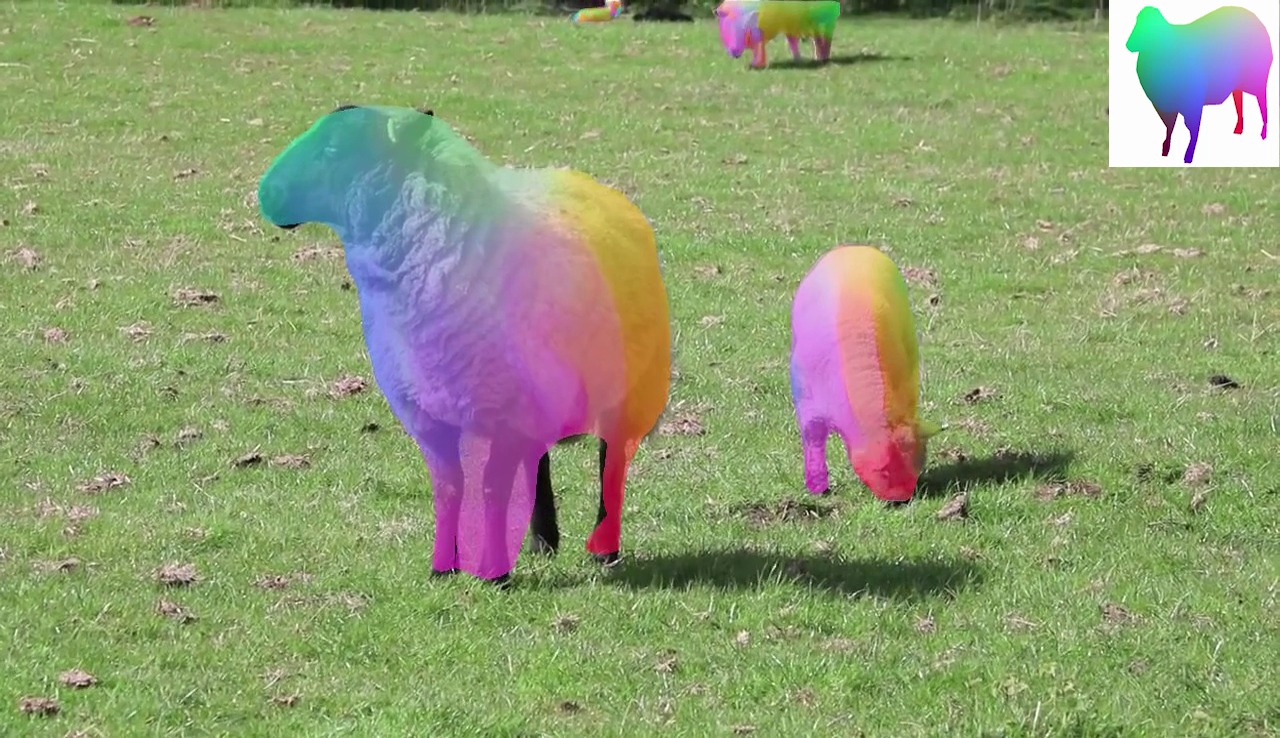} \\
\end{tabular}
}
\vspace{-0.1in}
\captionof{figure}{Snaps of a few frames from the Supplementary Video. We downloaded videos from youtube for 6 categories to show our results. We show the template shape in a canonical view on the top-right corner of the image.  A few of the car videos in the qualitative results were taken from CarFusion dataset  ~\cite{Narapureddy-2018-105893} \\\emph{Failure Modes} Our method has failure modes when the segmentation masks from Mask-RCNN \cite{he2017mask} are incorrect. Furthmore, since our method is trained on images with a single unoccluded/untruncated object per image hence our predictions are might be inaccurate for occluded objects or partially visible objects.}
\label{fig:video}
\end{table*} 
\section{Additional Result Visualization}
We show additional results on all the categories in Figure \ref{fig:add-bird}, \ref{fig:add-car}, \ref{fig:add-horse}, \ref{fig:add-zebra}, \ref{fig:add-cow}, \ref{fig:add-sheep}
\begin{table*}
\setlength{\tabcolsep}{0.05em}
\centering
  \scalebox{0.85}{
\begin{tabular}{cccccc}
\addpic{figures/images_st_bird_full/mb_2.jpg}  & 
\addpic{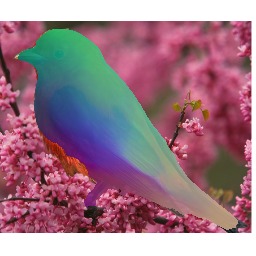}  & 
\addpic{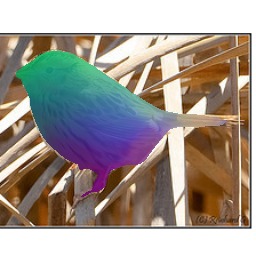}  & 
\addpic{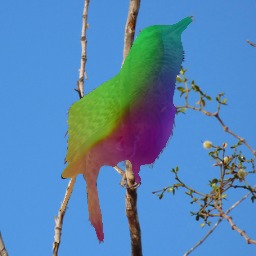}  & 
\addpic{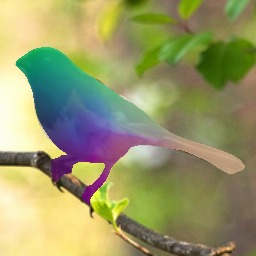}  & 
\addpic{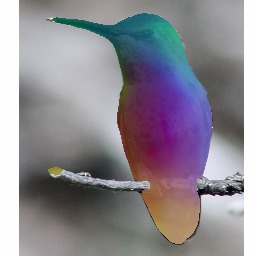}  \\ 
\addpic{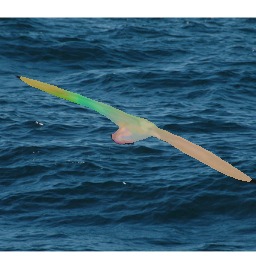}  & 
\addpic{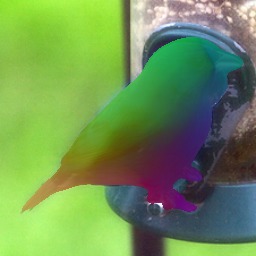}  & 
\addpic{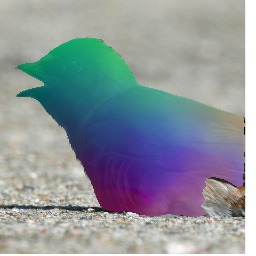}  & 
\addpic{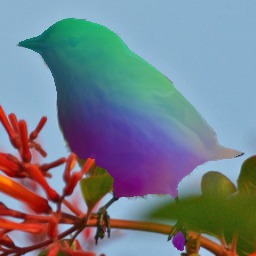}  & 
\addpic{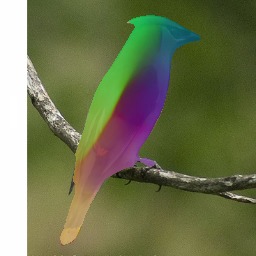}  & 
\addpic{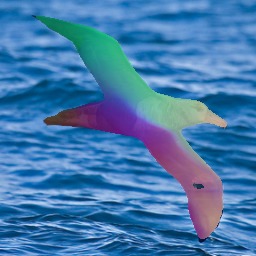}  \\ 
\addpic{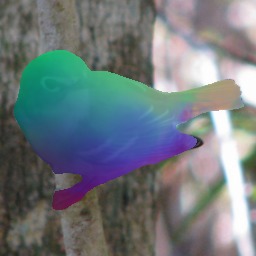}  & 
\addpic{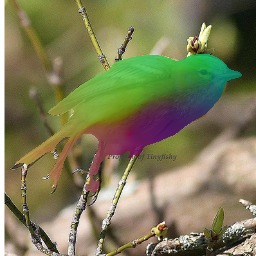}  & 
\addpic{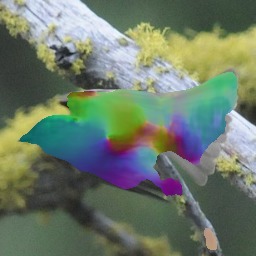}  & 
\addpic{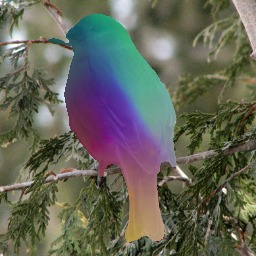}  & 
\addpic{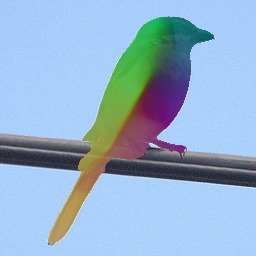}  & 
\addpic{figures/images_st_bird_full/bird_101.jpg}  \\ 
\end{tabular}
}
\vspace{-0.1in}
\captionof{figure}{Results of randomly sampled birds from the validation set}
\label{fig:add-bird}
\end{table*} 
\begin{table*}
\setlength{\tabcolsep}{0.05em}
\centering
  \scalebox{0.85}{
\begin{tabular}{cccccc}
\addpic{figures/images_st_car_full/mb_9.jpg}  & 
\addpic{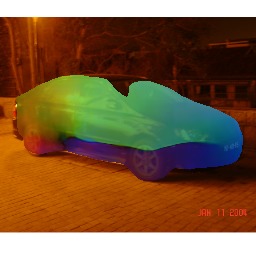}  & 
\addpic{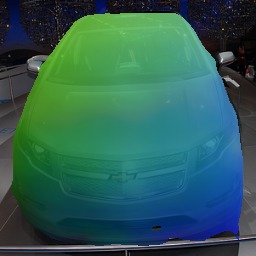}  & 
\addpic{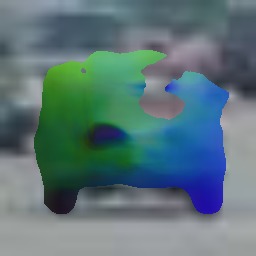}  & 
\addpic{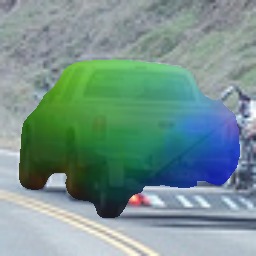}  & 
\addpic{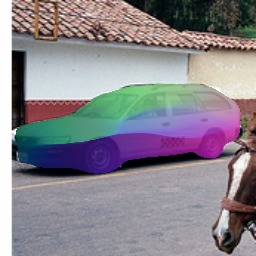}  \\ 
\addpic{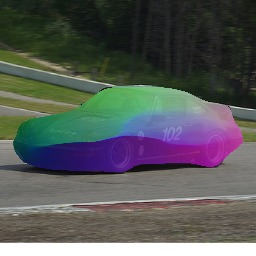}  & 
\addpic{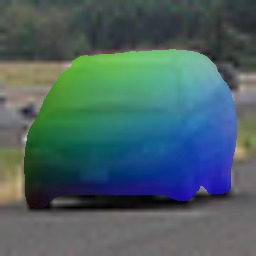}  & 
\addpic{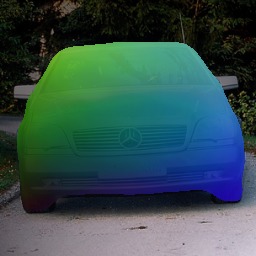}  & 
\addpic{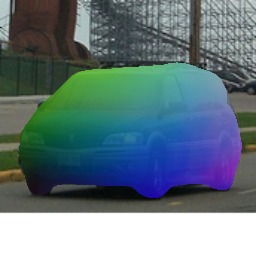}  & 
\addpic{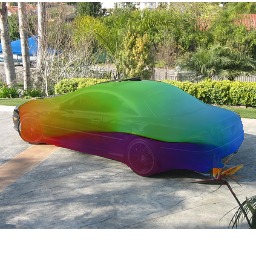}  & 
\addpic{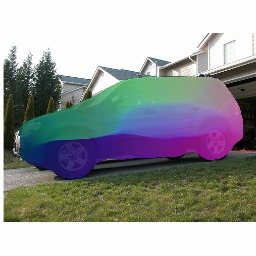}  \\ 
\addpic{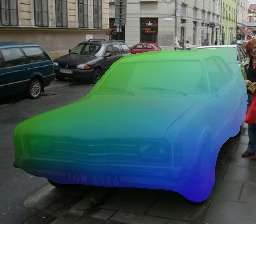}  & 
\addpic{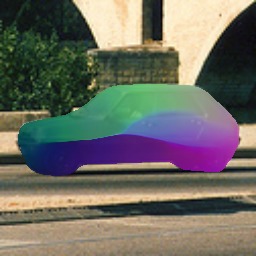}  & 
\addpic{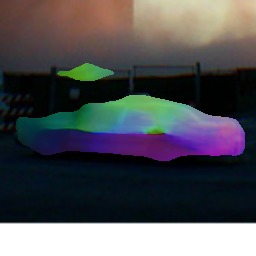}  & 
\addpic{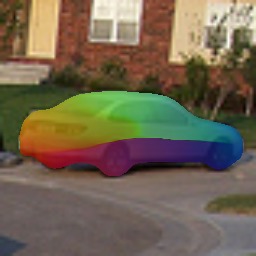}  & 
\addpic{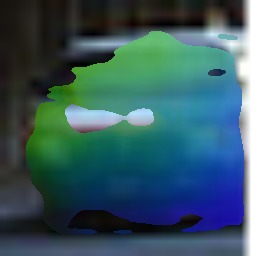}  & 
\addpic{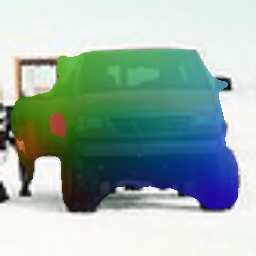}  \\ 
\end{tabular}
}
\vspace{-0.1in}
\captionof{figure}{Results of randomly sampled cars from the validation set}
\label{fig:add-car}
\end{table*} 
\begin{table*}
\setlength{\tabcolsep}{0.05em}
\centering
  \scalebox{0.85}{
\begin{tabular}{cccccc}
\addpic{figures/images_st_horse_full/mb_2.jpg}  & 
\addpic{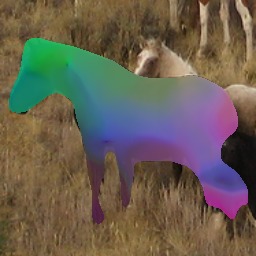}  & 
\addpic{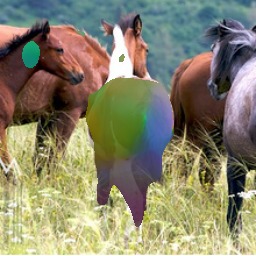}  & 
\addpic{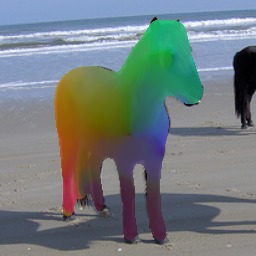}  & 
\addpic{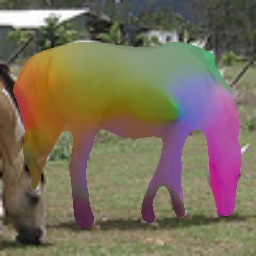}  & 
\addpic{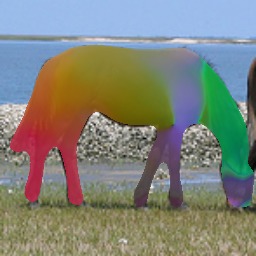}  \\ 
\addpic{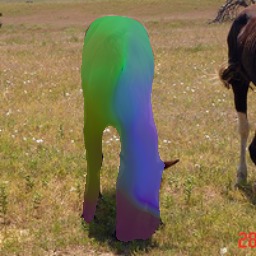}  & 
\addpic{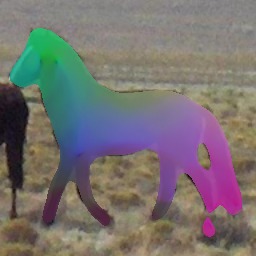}  & 
\addpic{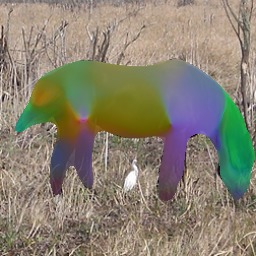}  & 
\addpic{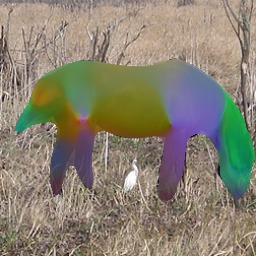}  & 
\addpic{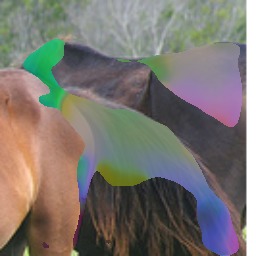}  & 
\addpic{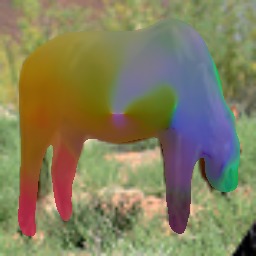}  \\ 
\addpic{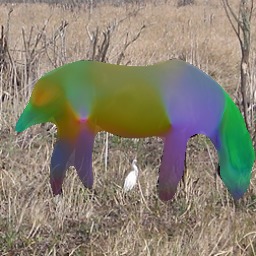}  & 
\addpic{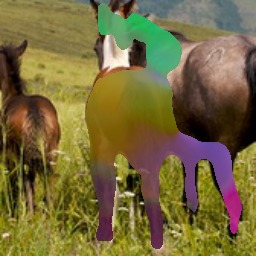}  & 
\addpic{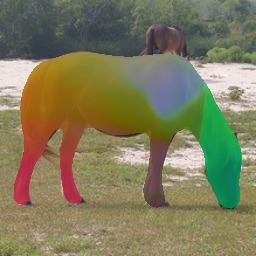}  & 
\addpic{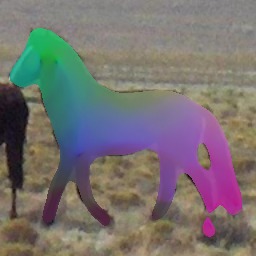}  & 
\addpic{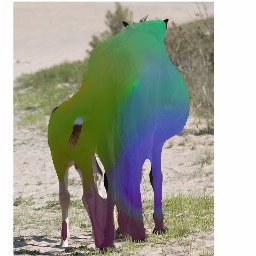}  & 
\addpic{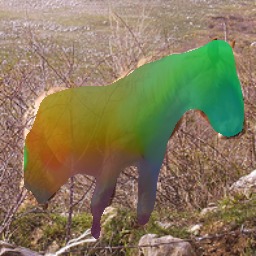}  \\ 
\end{tabular}
}
\vspace{-0.1in}
\captionof{figure}{Results of randomly sampled horses from the validation set}
\label{fig:add-horse}
\end{table*} 
\begin{table*}
\setlength{\tabcolsep}{0.05em}
\centering
  \scalebox{0.85}{
\begin{tabular}{cccccc}
\addpic{figures/images_st_zebra_full/mb_5.jpg}  & 
\addpic{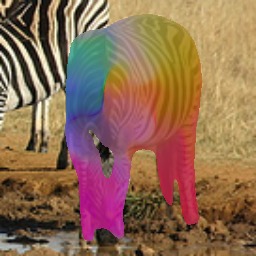}  & 
\addpic{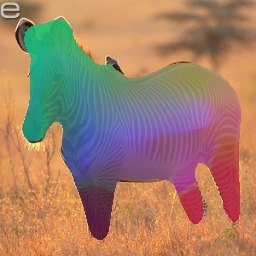}  & 
\addpic{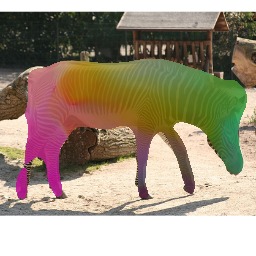}  & 
\addpic{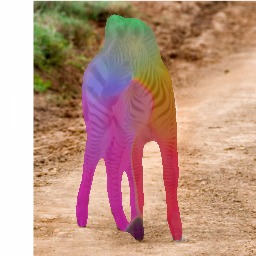}  & 
\addpic{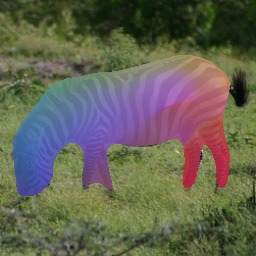}  \\ 
\addpic{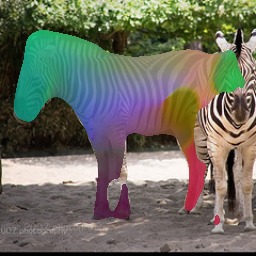}  & 
\addpic{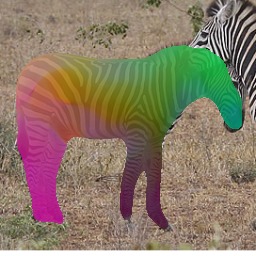}  & 
\addpic{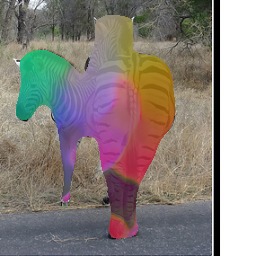}  & 
\addpic{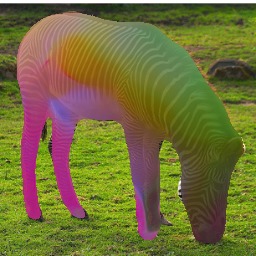}  & 
\addpic{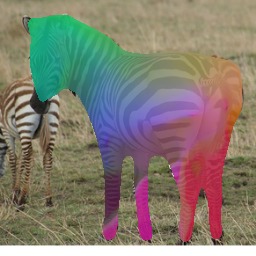}  & 
\addpic{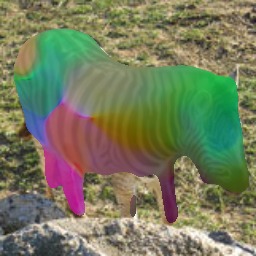}  \\ 
\addpic{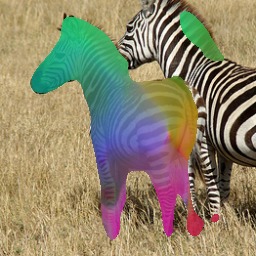}  & 
\addpic{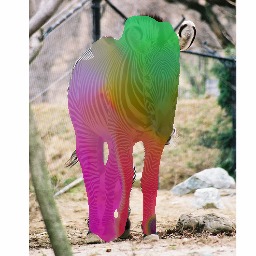}  & 
\addpic{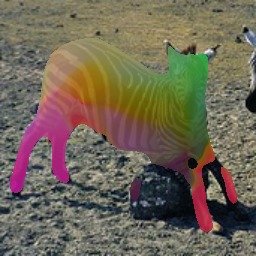}  & 
\addpic{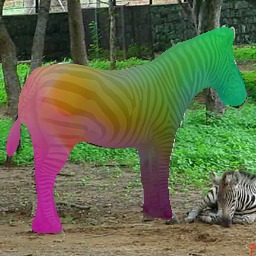}  & 
\addpic{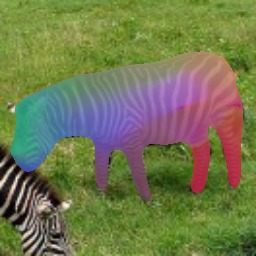}  & 
\addpic{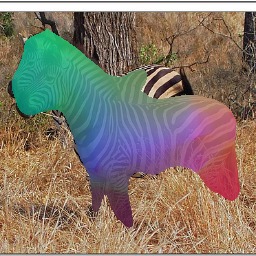}  \\
\end{tabular}
}
\vspace{-0.1in}
\captionof{figure}{Results of randomly sampled zebras from the validation set}
\label{fig:add-zebra}
\end{table*} 
\begin{table*}
\setlength{\tabcolsep}{0.05em}
\centering
  \scalebox{0.85}{
\begin{tabular}{cccccc}
\addpic{figures/images_st_cow_full/mb_0.jpg}  & 
\addpic{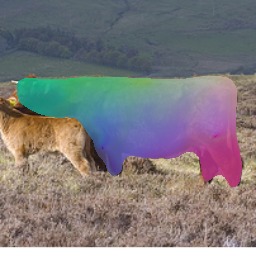}  & 
\addpic{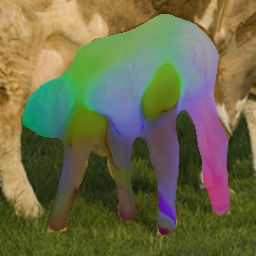}  & 
\addpic{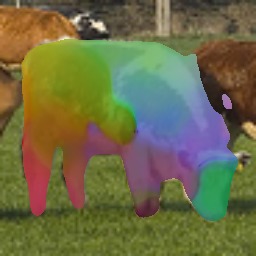}  & 
\addpic{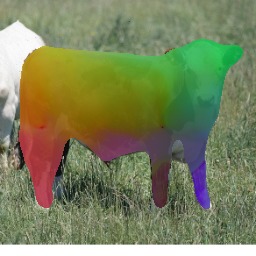}  & 
\addpic{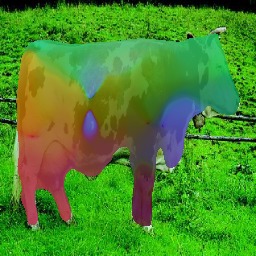}  \\ 
\addpic{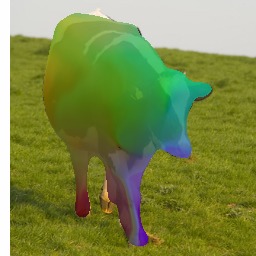}  & 
\addpic{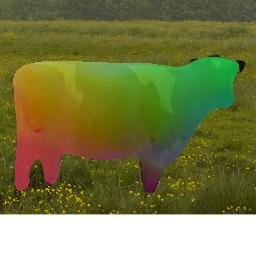}  & 
\addpic{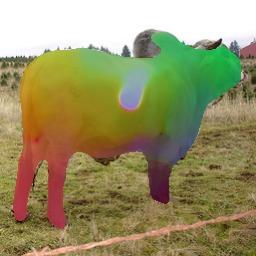}  & 
\addpic{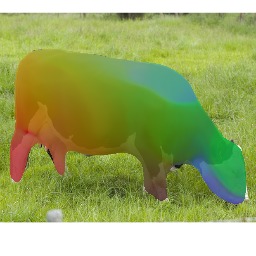}  & 
\addpic{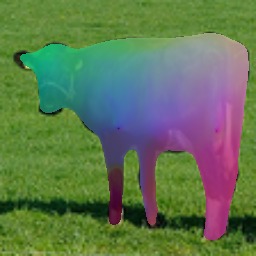}  & 
\addpic{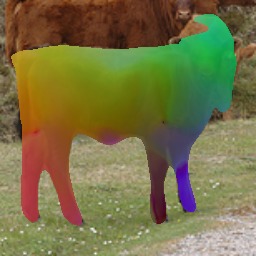}  \\ 
\addpic{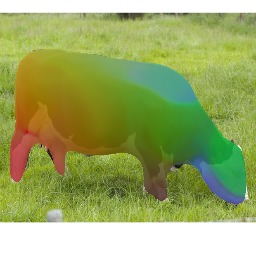}  & 
\addpic{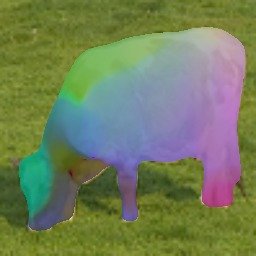}  & 
\addpic{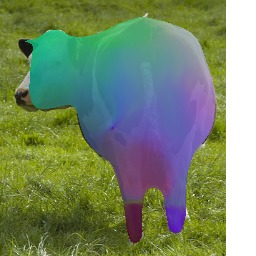}  & 
\addpic{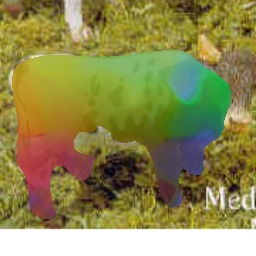}  & 
\addpic{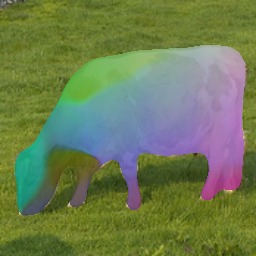}  & 
\addpic{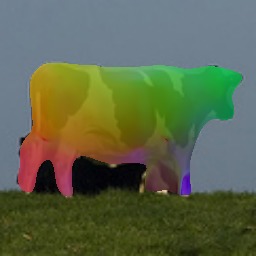}  \\
\end{tabular}
}
\vspace{-0.1in}
\captionof{figure}{Results of randomly sampled cows for the validation set}
\label{fig:add-cow}
\end{table*} 
\begin{table*}
\setlength{\tabcolsep}{0.05em}
\centering
  \scalebox{0.85}{
\begin{tabular}{cccccc} 
\addpic{figures/images_st_sheep_full/mb_2.jpg}  & 
\addpic{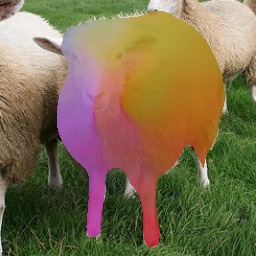}  & 
\addpic{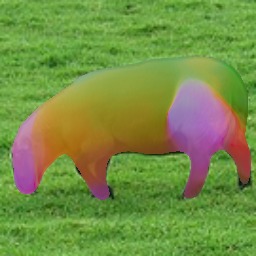}  & 
\addpic{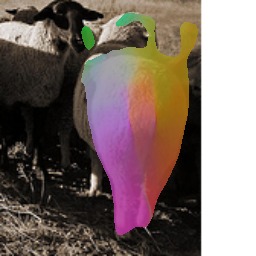}  & 
\addpic{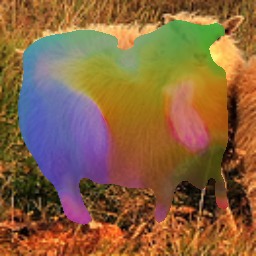}  & 
\addpic{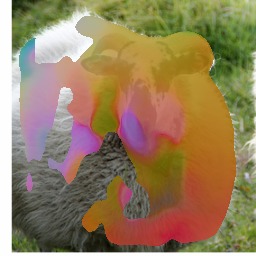}  \\ 
\addpic{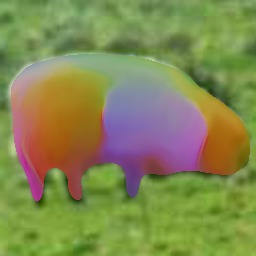}  & 
\addpic{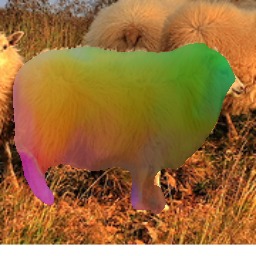}  & 
\addpic{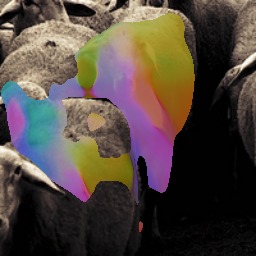}  & 
\addpic{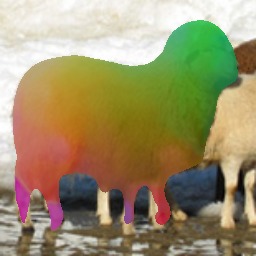}  & 
\addpic{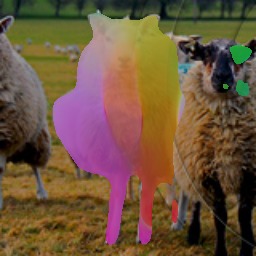}  & 
\addpic{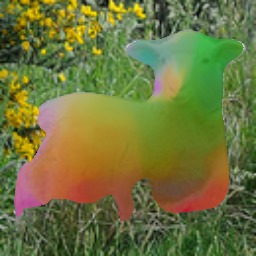}  \\ 
\addpic{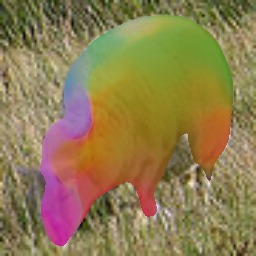}  & 
\addpic{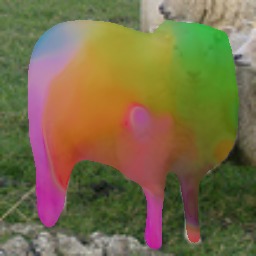}  & 
\addpic{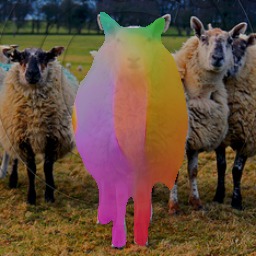}  & 
\addpic{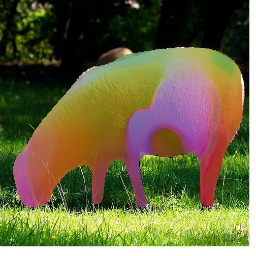}  & 
\addpic{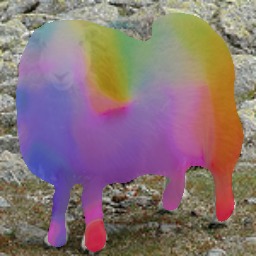}  & 
\addpic{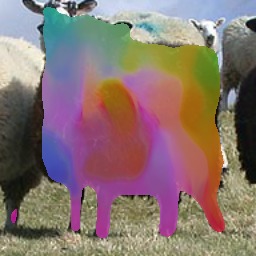}  \\ 
\end{tabular}
}
\vspace{-0.1in}
\captionof{figure}{Results of randomly sampled sheeps from the validation set }
\label{fig:add-sheep}
\end{table*} 

\end{document}